\documentclass[a4paper,review]{cas-sc}

\usepackage[numbers,sort&compress]{natbib}

\def\tsc#1{\csdef{#1}{\textsc{\lowercase{#1}}\xspace}}
\tsc{WGM}
\tsc{QE}
\tsc{EP}
\tsc{PMS}
\tsc{BEC}
\tsc{DE}
\usepackage{tikz}    
\usepackage{float}
\usepackage{bm}
\usepackage{subcaption}
\usepackage[justification=centering]{caption}
\usepackage{graphicx}%
\usepackage{multirow}%
\usepackage{amsmath,amssymb,amsfonts}%
\usepackage{amsthm}%

\makeatletter
\g@addto@macro\normalsize{%
  \setlength{\abovedisplayskip}{6\p@ \@plus 2\p@ \@minus 2\p@}%
  \setlength{\abovedisplayshortskip}{3\p@ \@plus 2\p@}%
  \setlength{\belowdisplayskip}{6\p@ \@plus 2\p@ \@minus 2\p@}%
  \setlength{\belowdisplayshortskip}{4\p@ \@plus 2\p@ \@minus 2\p@}%
}
\makeatother
\makeatletter
\def\@listi{\leftmargin\leftmargini
  \parsep    0\p@ \@plus 1\p@ \@minus 1\p@
  \topsep    4\p@ \@plus 2\p@ \@minus 2\p@
  \itemsep   2\p@ \@plus 1\p@ \@minus 1\p@}
\let\@listI\@listi  % keep this as the value restored when \normalsize is reselected
\@listi             % apply immediately (the original cas defaults already ran)
\makeatother
\AtBeginEnvironment{align}{}
\AtBeginEnvironment{align*}{}
\AtBeginEnvironment{gather}{}
\AtBeginEnvironment{gather*}{}
\AtBeginEnvironment{multline}{}
\AtBeginEnvironment{multline*}{}
\AtBeginEnvironment{eqnarray}{}
\AtBeginEnvironment{eqnarray*}{}
\allowdisplaybreaks[1]
\usepackage[title]{appendix}%
\usepackage{xcolor}%
\usepackage{textcomp}%
\usepackage{manyfoot}
\usepackage{booktabs}%
\usepackage{algorithm}%
\usepackage{algorithmicx}%
\usepackage{algpseudocode}%
\usepackage{listings}%
\usepackage{bbding}
\usepackage{wrapfig}
\usepackage{placeins}
\begin{document}
\let\WriteBookmarks\relax
% Float-placement tuning so that figures stay near the text that
% references them (instead of being deferred to the end of the document).
% The earlier \def\floatpagepagefraction / \def\textpagefraction lines
% were removed: those control sequences do not exist in LaTeX (the real
% parameters \floatpagefraction and \textfraction are already configured
% in cas-common.sty), so the lines were inert typos that obscured intent.
\renewcommand{\topfraction}{0.95}
\renewcommand{\bottomfraction}{0.95}
\renewcommand{\textfraction}{0.05}
\renewcommand{\floatpagefraction}{0.5}
\setlength{\textfloatsep}{8pt plus 2pt minus 2pt}
\setlength{\floatsep}{8pt plus 2pt minus 2pt}
\setlength{\intextsep}{8pt plus 2pt minus 2pt}
\setcounter{topnumber}{4}
\setcounter{bottomnumber}{4}
\setcounter{totalnumber}{8}
\shorttitle{Categorical Optimization via COBALT}
\shortauthors{Z.Y. Liang et~al.}
%\begin{frontmatter}

\title [mode = title]{Categorical optimization via Bayesian anchored latent trust-regions for structural design under high-dimensional uncertainties}                      
\tnotemark[1,2]

\author[1]{Zhangyong Liang}

\author[2]{Jie Hou}

\author[3]{Huanhuan Gao}[orcid=0000-0003-4463-6433]
\cormark[1]
\ead{gao_huanhuan@jlu.edu.cn}

\author[4]{Manyu Xiao}
\cormark[2]
\ead{manyuxiao@nwpu.edu.cn}

\affiliation[1]{organization={National Center for Applied Mathematics, Tianjin University},
                addressline={Weijin Road 92},
                postcode={300072},
                city={Tianjin},
                country={China}}

\affiliation[2]{organization={School of Mechanical Engineering, Northwestern Polytechnical University},
                addressline={Youyixi Road 127},
                postcode={710072},
                city={Xi'an},
                country={China}}

\affiliation[3]{organization={School of Mechanical and Aerospace Engineering, Jilin University},
                addressline={Renmin Street 5988},
                postcode={130025},
                city={Changchun},
                country={China}}

\affiliation[4]{organization={Xi'an Key Laboratory of Scientific Computation and Applied Statistics, School of Mathematics and Statistics, Northwestern Polytechnical University},
                addressline={Youyixi Road 127},
                postcode={710072},
                city={Xi'an},
                country={China}}

\cortext[cor1]{Corresponding author}
\cortext[cor2]{Corresponding author}

\begin{abstract}
  Categorical structural optimization under aleatoric uncertainties presents significant challenges due to the requirement of selecting each design variable from a finite catalog of allowable instances. Moreover, each candidate design necessitates costly stochastic finite-element evaluations. Existing latent-space optimization methods, while capable of reducing the dimensionality of catalog attributes, typically treat the resultant space as a continuous search domain. This approach necessitates rounding off the continuous optimum to the nearest catalog instance, potentially altering the objective value, constraint status, or physical interpretation of the design. To mitigate this issue, this paper introduces the Categorical Optimization with Bayesian Anchored Latent Trust-Regions (COBALT) framework, designed for high-dimensional categorical optimization under uncertainties (OUU). COBALT initially embeds the physical catalog into a low-dimensional latent representation, locking the mapped instances as a discrete anchored graph. Subsequently, a data-independent random tree decomposition is employed to offer bounded-complexity additive modeling for high-dimensional categorical variables. On this anchored domain, an additive Sparse Axis-Aligned Subspace Gaussian Process (SAAS-GP) surrogate is fitted to heteroscedastic Monte Carlo-based Finite Element Analysis (MC-FEA) observations. A trust-region discrete graph acquisition search is then used to select the next admissible catalog configuration, avoiding continuous relaxation or rounding-off. The proposed strategy is demonstrated in the robust design optimization of complex beam structures, considering factors such as structural weight, strain energy, and local buckling performance. Numerical examples indicate that evaluating only valid catalog designs through the MC-FEA oracle preserves physical admissibility throughout the active learning loop. Furthermore, it can enhance the efficiency of robust categorical structural optimization when compared to relaxation-based and non-surrogate search strategies under equivalent conditions.
\end{abstract}

% \begin{graphicalabstract}
% \includegraphics{figs/cas-grabs.pdf}
% \end{graphicalabstract}

\begin{highlights}
  \item COBALT anchors discrete latent instances to avoid infeasible decoding in search.
  \item COBALT trust-region graph search selects valid catalog designs without rounding-off.
  \item Random tree decomposition limits additive surrogate complexity in COBALT.
  \item Fully Bayesian sparse surrogate inference models noisy stochastic simulation data.
  \item Structural benchmarks demonstrate performance from grouped to high-dimensional catalogs.
\end{highlights}


\maketitle

\section{Introduction}
\label{sc1}
In various optimization problems, design variables are generally classified as continuous, discrete (including integer), and categorical ones \cite{kokkolaras2001mixed,lindroth2011pure}. Unlike continuous or integer variables, a categorical variable takes values from a finite set of predefined instances, rather than arbitrary real numbers \cite{coelho2015investigation,sloane1996introduction}. A typical structural example is the selection of a beam cross-section from a catalog of standard profiles. These catalog sections may differ simultaneously in shape, area and moments of inertia, whcih cannot be faithfully represented by a single scalar index.

Classical methods for categorical optimization problems often transform available instances into artificial real-valued codes or binary strings \cite{caruana1988representation,eshelman1993real}. Evolutionary optimizers such as genetic algorithms are then applied to search the encoded space \cite{herrera1998tackling,goldberg2006genetic,liao2014ant,rajeev1992discrete}. However, the numerical distance between two encoded values usually has no direct relation to the physical difference between the corresponding catalog instances, which may reduce the search efficiency and obscure the interpretation of the optimized structural design.

Other works have been proposed to treat categorical variables in a more specific manner \cite{herrera2014metamodel,filomeno2012extending,mccane2008distance}. For example, when dealing with unordered categorical variables, simplex coding is broadly applied, which suits evolutionary methods well \cite{coelho2015investigation,filomeno2014metamodels,fu2008classification}. In this coding, the distances between any two design instances are kept equal, thus forming a regular simplex in the space. However, if the number of instances is large, the dimensionality of the related space becomes computationally prohibitive. The work \cite{WOS:000522857300001} proposes a bi-level optimization framework for large-scale mixed categorical structural optimization problems, in which the problems are divided into continuous slave problems and discrete master problems. The subsequent extension can also be found in \cite{WOS:000836612800008}.

Based on this multi-dimensional representation, manifold learning techniques can be utilized to discover the lower-dimensional intrinsic manifold embedded in the higher-dimensional attribute space of the catalog. The main linear manifold learning techniques are Principal Component Analysis (PCA) \cite{jolliffe2002principal} and Multi-Dimensional Scaling (MDS) \cite{martin1993measuring}. PCA applies eigenvalue decomposition to preserve the most covariance information in the reduced-order space. MDS instead focuses on preserving Euclidean distances between sample points. However, they lack the ability to map a non-linear manifold to a reduced-order space \cite{tenenbaum2000global}. Typical non-linear manifold learning techniques include Locally Linear Embedding (LLE) \cite{roweis2000nonlinear,de2003supervised}, Kernel Principal Component Analysis (KPCA) \cite{scholkopf1998nonlinear,cao2003comparison}, and Isomap \cite{tenenbaum2000global,balasubramanian2002isomap}. Particularly, Isomap is a variant of MDS that replaces Euclidean distances with geodesic distances computed by the Dijkstra algorithm \cite{dijkstra1959note}. 

In previous deterministic optimization frameworks, the reduced-order representation was further approximated by continuous polynomial interpolations. This allowed gradient-based methods, such as the Method of Moving Asymptotes (MMA) \cite{svanberg1987method,svanberg1993method}, to be applied on a continuous manifold. Since the original categorical problem admits only finite catalog instances, the continuous optimum must then be rounded off to a nearby admissible instance through a nearest-neighbor search \cite{shakhnarovich2006nearest}. However, the rounded design may differ from the continuous solution in objective value, constraint status, and physical interpretation. We refer to this mismatch as decoding failure (or Rounding-off Error), which becomes more pronounced when the catalog manifold is highly non-linear or sparsely sampled.

The above difficulty is further amplified when structural design is performed under aleatoric uncertainties, such as manufacturing tolerances, material property fluctuations, and environmental load variations. Transitioning from deterministic categorical optimization to Optimization Under uncertainties (OUU) requires evaluating robust performance metrics through stochastic simulations, for example Monte Carlo-based Finite Element Analysis (MC-FEA). Each admissible catalog combination is then associated with expensive statistical response quantities rather than a single deterministic response, and the resulting observations are generally noisy and heteroscedastic. This makes direct enumeration, deterministic continuous optimizers, and traditional heuristic global searches \cite{abramson2009mesh,stegmann2005discrete} inefficient for high-dimensional categorical OUU.

Bayesian Optimization (BO) \cite{7352306,2018arXiv180702811F} provides a principled surrogate-assisted framework for expensive and noisy evaluations. By maintaining a probabilistic surrogate model, typically a Gaussian Process (GP) \cite{NIPS1995_7cce53cf,Rasmussen2006Gaussian}, BO balances exploration and exploitation while propagating predictive uncertainties throughout the optimization loop \cite{NIPS2012_05311655}. 
Nevertheless, standard BO scales poorly with high-dimensional combinatorial spaces. 
Dimensionality-reduction-enhanced BO methods (i.e., Latent Space BO) have been proposed to map high-dimensional design instances onto a lower-dimensional latent representation \cite{wan2021think,Deshwal2023BODi}. 
However, existing frameworks often optimize a continuous acquisition function in the latent space. 
Consequently, when the feasible designs must remain strict catalog instances, these methods still require a decoding or rounding-off step and remain vulnerable to the mismatch described above. 
The paper \cite{van2025parallel} proposes a robust decomposition-based approach combining parallel constrained Bayesian optimization with active learning-based reliability evaluation. The work \cite{schneider2024maximum} presents a Bayesian optimization framework for MAP estimation in structural dynamics. Bootstrapped neural ensembles with randomized priors have also been used for Bayesian optimization of high-dimensional black-box functions, which has shown strong performance in challenging applications \cite{bhouri2023scalable}. 
The study \cite{hong2023collaborative} extends uncertainties quantification capabilities to all three types of probabilistic models and can bound measures such as output variance and failure probability. 
The paper \cite{kaczmarski2023bayesian} proposes a generalized Bayesian optimization method for fiber-based biomimetic soft-robotic arms which can meet robotic control requirements through computationally robust reduced-order active filament modeling.

To address the decoding failure and dimensionality challenges in high-dimensional categorical OUU, this paper proposes the \textbf{C}ategorical \textbf{O}ptimization with \textbf{B}ayesian \textbf{A}nchored \textbf{L}atent \textbf{T}rust-Regions (\textbf{COBALT}) framework. This framework integrates discrete manifold anchoring with Bayesian optimization under uncertainties. The main contributions of this work are summarized as follows:
\begin{itemize}
    \item \textbf{From continuous relaxation to absolute discrete anchoring.} COBALT does not treat the dimensionality-reduced latent space as a searchable continuum. It instead locks the mapped catalog instances as a fixed network of admissible anchors. The search therefore stays on valid categorical designs throughout.
    \item \textbf{Data-independent random tree decomposition.} COBALT avoids learning a fragile interaction structure from sparse, noisy observations. It samples random tree decompositions over the categorical variables. The decompositions give a bounded-complexity view of high-dimensional catalog combinations and vary the pairwise interactions seen by the surrogate from iteration to iteration.
    \item \textbf{uncertainties-aware sparse inference.} Multi-component catalog selection causes rapid combinatorial growth. MC-FEA further contributes heteroscedastic observation noise. We tackle both with Sparse Axis-Aligned Subspace (SAAS) priors on the Gaussian Process. Under fully Bayesian inference, the surrogate isolates influential latent features and interactions while staying robust to noise.
    \item \textbf{Discrete graph acquisition search.} COBALT does not optimize a continuous acquisition function and then round off to the catalog. Instead, Dijkstra-based graph-evolutionary operators run inside the Bayesian active learning loop. The search traverses the discrete anchored graph within dynamically scaled latent trust-regions. The recommended design is therefore a physically admissible configuration before MC-FEA is called.
    \item \textbf{Robust MC-FEA evaluation of valid catalog designs.} COBALT separates deterministic catalog geometry from aleatoric uncertainties. The geometry is fixed by the anchored latent graph. Aleatoric uncertainties is then evaluated only on an admissible categorical configuration. The MC-FEA oracle thus estimates robust objectives and checks problem-specific constraints without further decoding ambiguity.
\end{itemize}

The paper is organized as follows. Section~\ref{sc1} sets out the background and motivation. Section~\ref{sc2} formulates the robust categorical structural optimization problem under uncertainties. Section~\ref{sc3} develops the COBALT framework: anchored discrete latent manifolds, data-independent random decompositions, SAAS-GP surrogate modeling, and trust-region discrete graph acquisition. Section~\ref{sc4} reports numerical tests on complex structures under uncertainties. Section~\ref{sc5} concludes and outlines future work.

\section{Problem formulation}
\label{sc2}

\subsection{Categorical structural optimization under uncertainties}
\label{sc2_1}
In categorical structural optimization, the admissible instances usually correspond to available engineering choices such as standard cross-section types selected from a design catalog.
Each instance is described by a multi-dimensional discrete vector, and the components of this vector are the physical attributes of the corresponding design choice.
Therefore, an available catalog can be represented as a finite set of points in a multi-dimensional attribute space.

Let $e$ denote the number of categorical design variables.
For the $i$-th categorical variable, its admissible catalog is denoted by
\begin{equation}
    \textbf{X}_{\mbox{\tiny{i}}}=\{\textbf{x}_{\mbox{\tiny{i}}}^{\mbox{\tiny{1}}},\textbf{x}_{\mbox{\tiny{i}}}^{\mbox{\tiny{2}}},\cdots,\textbf{x}_{\mbox{\tiny{i}}}^{\mbox{\tiny{{n}}}}\},\quad \textbf{x}_{\mbox{\tiny{i}}}\in\textbf{X}_{\mbox{\tiny{i}}},\quad i=1,2,\cdots,e.
\end{equation}

The selected value can equivalently be indicated by a catalog label $c_i\in\{1,2,\cdots,n\}$ with $\textbf{x}_{\mbox{\tiny{i}}}=\textbf{x}_{\mbox{\tiny{i}}}^{\mbox{\tiny{$c_i$}}}$. 
Each instance is an $M$-dimensional nominal attribute vector,
\begin{equation}
    \textbf{x}_{\mbox{\tiny{i}}}^{\mbox{\tiny{j}}}=({}^{\mbox{\tiny{1}}}a_{\mbox{\tiny{i}}}^{\mbox{\tiny{j}}},{}^{\mbox{\tiny{2}}}a_{\mbox{\tiny{i}}}^{\mbox{\tiny{j}}},\cdots,{}^{\mbox{\tiny{M}}}a_{\mbox{\tiny{i}}}^{\mbox{\tiny{j}}})^{\mbox{\tiny{T}}},
    \quad j=1,2,\cdots,n.
\end{equation}

Thus, a complete categorical design is written as $\textbf{x}_{1:e}=(\textbf{x}_{\mbox{\tiny{1}}},\textbf{x}_{\mbox{\tiny{2}}},\cdots,\textbf{x}_{\mbox{\tiny{e}}})\in\mathcal{X}=\textbf{X}_{\mbox{\tiny{1}}}\times\textbf{X}_{\mbox{\tiny{2}}}\times\cdots\times\textbf{X}_{\mbox{\tiny{e}}}$, where $\mathcal{X}$ is a finite categorical design space containing $n^e$ possible combinations.

\begin{figure}
    \centering
    \includegraphics[width=0.45\columnwidth]{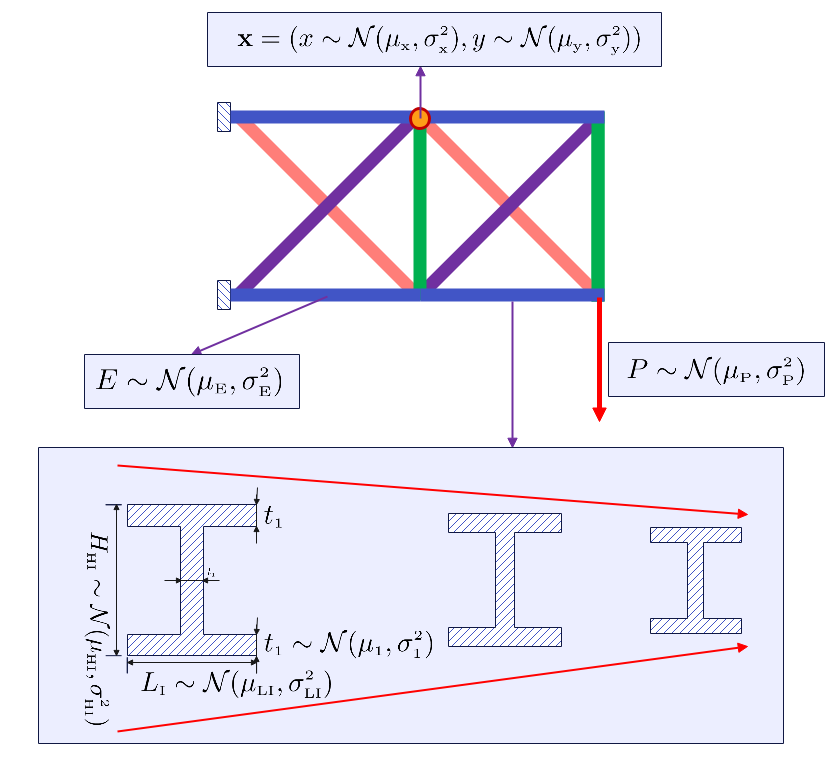}
    \caption{The structural uncertainties introduced by material properties, external loads, geometric tolerance, and cross section shape errors.}
    \label{fig:Ibeam}
\end{figure}
In real-world structural systems are subjected to aleatoric uncertainties, such as manufacturing tolerances, material property fluctuations, and unpredictable environmental loads, as shown in Fig. \ref{fig:Ibeam}. Even for a fixed I-beam instance selected from the catalog, its geometric attributes such as flange width, flange thickness, and section height may fluctuate around their nominal values. As a result, the deterministic categorical optimization problem must be extended to OUU. 
Let $\boldsymbol{\xi}\in\Xi$ denote the random vector collecting all aleatoric sources, with probability law $P_{\boldsymbol{\xi}}$.
For a fixed categorical design, the structural model assembled from the selected catalog instances is analyzed under different realizations of $\boldsymbol{\xi}$.
For consistency with the numerical examples, the robust categorical structural optimization problem with $e$ categorical variables and $q$ scalar feasibility margins is written as follows:
\begin{equation}
    \label{opt_sta1}
    \begin{split}
        &\text{min.:}\hspace{5mm}\mathcal{J}_{\texttt{robust}}(\textbf{x}_{\mbox{\tiny{1}}}, \textbf{x}_{\mbox{\tiny{2}}}, \cdots, \textbf{x}_{\mbox{\tiny{e}}}) = \mathbb{E}_{\boldsymbol{\xi}}[J(\textbf{x}_{\mbox{\tiny{1}}}, \textbf{x}_{\mbox{\tiny{2}}}, \cdots, \textbf{x}_{\mbox{\tiny{e}}}, \boldsymbol{\xi})] + \gamma \sqrt{\mathbb{V}_{\boldsymbol{\xi}}[J(\textbf{x}_{\mbox{\tiny{1}}}, \textbf{x}_{\mbox{\tiny{2}}}, \cdots, \textbf{x}_{\mbox{\tiny{e}}}, \boldsymbol{\xi})]};\\
        &\text{s.}\hspace{1.5mm}\text{t.:}\hspace{5.8mm}\textbf{h}(\textbf{x}_{\mbox{\tiny{1}}}, \textbf{x}_{\mbox{\tiny{2}}}, \cdots, \textbf{x}_{\mbox{\tiny{e}}}) \leq \textbf{0};\\
        &\hspace{13mm}\textbf{v}(\textbf{x}_{\mbox{\tiny{1}}}, \textbf{x}_{\mbox{\tiny{2}}}, \cdots, \textbf{x}_{\mbox{\tiny{e}}}) = \textbf{0};\\
        &\hspace{13mm}\textbf{x}_{\mbox{\tiny{i}}}\in\textbf{X}_{\mbox{\tiny{i}}}=\{\textbf{x}_{\mbox{\tiny{i}}}^{\mbox{\tiny{1}}},\textbf{x}_{\mbox{\tiny{i}}}^{\mbox{\tiny{2}}},\cdots,\textbf{x}_{\mbox{\tiny{i}}}^{\mbox{\tiny{{n}}}}\},\hspace{2mm}\mbox{\normalsize{i}}=1,2,\cdots,{\mbox{\normalsize{{e}}}};\\
        &\hspace{13mm} \textbf{x}_{\mbox{\tiny{i}}}^{\mbox{\tiny{j}}}=({}^{\mbox{\tiny{1}}}a_{\mbox{\tiny{i}}}^{\mbox{\tiny{j}}},{}^{\mbox{\tiny{2}}}a_{\mbox{\tiny{i}}}^{\mbox{\tiny{j}}},\cdots,{}^{\mbox{\tiny{M}}}a_{\mbox{\tiny{i}}}^{\mbox{\tiny{j}}})^{\mbox{\tiny{T}}},\hspace{2mm}\mbox{\normalsize{j}}=1,2,\cdots,{\mbox{\normalsize{{n}}}}.\\
    \end{split}
\end{equation}

In Eq.~\eqref{opt_sta1}, $J(\cdot,\boldsymbol{\xi})$ is the scalar structural response (e.g., strain energy) from one finite-element realization under a sampled uncertainty vector. $\mathcal{J}_{\texttt{robust}}$ combines its expectation $\mathbb{E}_{\boldsymbol{\xi}}[\cdot]$ and standard deviation $\sqrt{\mathbb{V}_{\boldsymbol{\xi}}[\cdot]}$ with weight $\gamma \geq 0$ penalizing performance fluctuation. The inequality vector $\textbf{h}=(h_1,\ldots,h_q)^{\mbox{\tiny{T}}}$ collects benchmark-specific feasibility margins (e.g., mass and local buckling), with $h_r\leq 0$ meaning satisfied; $\textbf{v}=(v_1,\ldots,v_d)^{\mbox{\tiny{T}}}$ denotes the equality constraints. The $i$-th categorical design variable $\textbf{x}_{\mbox{\tiny{i}}}$ takes values from the finite instance set $\textbf{X}_{\mbox{\tiny{i}}}$. 
Its $j$-th catalog instance $\textbf{x}_{\mbox{\tiny{i}}}^{\mbox{\tiny{j}}}$ is an $M$-dimensional nominal attribute vector. Each entry ${}^{\mbox{\tiny{l}}}a_{\mbox{\tiny{i}}}^{\mbox{\tiny{j}}}$ is the $l$-th attribute (e.g., area, moment of inertia, torsional constant, or section modulus) required by the finite-element model.

For any fixed categorical combination $(\textbf{x}_{\mbox{\tiny{1}}}, \textbf{x}_{\mbox{\tiny{2}}}, \cdots, \textbf{x}_{\mbox{\tiny{e}}})$, the objective statistics in Eq. \eqref{opt_sta1} are estimated by repeatedly evaluating the assembled structural model under sampled uncertainties.
This evaluation relies on MC-FEA in the present work.
Because only a finite number $N_{MC}$ of samples can be used, the exact robust objective $\mathcal{J}_{\texttt{robust}}$ is not observed directly.
Instead, the optimizer receives an expensive noisy Monte Carlo estimator $y_{\texttt{obs}}$:
\begin{equation}
    \label{eq_noise}
    y_{\texttt{obs}}(\textbf{x}_{\mbox{\tiny{1}}}, \textbf{x}_{\mbox{\tiny{2}}}, \cdots, \textbf{x}_{\mbox{\tiny{e}}}) = \mathcal{J}_{\texttt{robust}}(\textbf{x}_{\mbox{\tiny{1}}}, \textbf{x}_{\mbox{\tiny{2}}}, \cdots, \textbf{x}_{\mbox{\tiny{e}}}) + \epsilon(\textbf{x}_{\mbox{\tiny{1}}}, \textbf{x}_{\mbox{\tiny{2}}}, \cdots, \textbf{x}_{\mbox{\tiny{e}}}), \quad \epsilon(\cdot) \sim \mathcal{N}(0, \sigma^2_{\epsilon}(\cdot)),
\end{equation}
where $\epsilon(\cdot)$ denotes the finite-sample Monte Carlo error and $\sigma^2_{\epsilon}(\cdot)$ is generally input-dependent, giving rise to heteroscedastic observations as modeled again in Eq.~\eqref{noise_model}.
More explicitly, for $N_{MC}$ independent samples $\{\boldsymbol{\xi}^{(r)}\}_{r=1}^{N_{MC}}$, the oracle computes
\begin{equation}
    \label{mc_estimator}
    \widehat{\mathcal{J}}_{\texttt{robust}}
    =
    \frac{1}{N_{MC}}\sum_{r=1}^{N_{MC}} J^{(r)}
    + \gamma
    \sqrt{
    \frac{1}{N_{MC}-1}\sum_{r=1}^{N_{MC}}
    \left(J^{(r)}-\frac{1}{N_{MC}}\sum_{\ell=1}^{N_{MC}}J^{(\ell)}\right)^2
    },
    \quad
    J^{(r)}=J(\textbf{x}_{\mbox{\tiny{1}}},\ldots,\textbf{x}_{\mbox{\tiny{e}}},\boldsymbol{\xi}^{(r)}).
\end{equation}

This sample estimator is the quantity passed to the optimizer as $y_{\texttt{obs}}$, while the sampling variability across different catalog combinations is absorbed into the heteroscedastic noise model.
The problem-specific feasibility margins $\textbf{h}$ are evaluated for each selected catalog design according to the corresponding benchmark mechanics. 

\subsection{Bayesian optimization}
\label{sc2_2}
In this paper, BO guides the search for a robust categorical design under a limited MC-FEA budget. 
As formalized in Section~\ref{sc3}, each admissible catalog combination has a one-to-one latent representation $\textbf{Z}\in\Omega_D\subset\mathbb{R}^D$. This $\textbf{Z}$ is not an independent continuous variable, only a numerical coordinate of a physically valid catalog combination. 
We therefore define the BO objective as $f(\textbf{Z})=\mathcal{J}_{\texttt{robust}}(\textbf{x}_{1:e}(\textbf{Z}))$, where $\textbf{x}_{1:e}=(\textbf{x}_{1},\ldots,\textbf{x}_{e})$ is the catalog combination uniquely represented by $\textbf{Z}$.
BO then minimizes $f$ sequentially, alternating between fitting a probabilistic surrogate to evaluated structures and selecting the next admissible candidate via an acquisition function.

\textbf{Gaussian process surrogate:} 
To approximate the robust structural response over catalog combinations with few MC-FEA evaluations, we use a GP surrogate. 
Following \cite{williams2006gaussian}, a GP can be interpreted from a weight-space view.
Given a set of $t$ evaluated admissible designs, denoted by $\mathcal{D}_t = \{ (\textbf{Z}^{(i)}, y_{\texttt{obs}}^{(i)}) \mid i=1,\ldots,t \}$, we build the surrogate as follows.
We first consider a linear surrogate:
\begin{equation}
    f(\textbf{Z}) = \phi(\textbf{Z})^{\mbox{\tiny{T}}} \textbf{w}, \quad y_{\texttt{obs}}=f(\textbf{Z}) + \epsilon(\textbf{Z}), \quad \epsilon(\textbf{Z}) \sim \mathcal{N}(0, \sigma_{\epsilon}^2(\textbf{Z})),
\end{equation}
where $\textbf{w}$ is the model parameter vector, $\phi(\textbf{Z})$ is the feature mapping, and $\epsilon(\textbf{Z})$ is the MC-FEA observation noise at the anchored design $\textbf{Z}$.
A zero-mean Gaussian prior $\textbf{w} \sim \mathcal{N}(\textbf{0}, \boldsymbol{\Sigma}_p)$ is placed on the weights, yielding an analytic predictive distribution. The feature mappings always appear as a $\boldsymbol{\Sigma}_p$-weighted inner product, comparing inputs in feature space and admitting the kernel trick. 
We thus replace the explicit feature map $\phi(\cdot)$ with a covariance kernel $k(\textbf{Z}, \textbf{Z}^\prime) = \phi(\textbf{Z})^{\mbox{\tiny{T}}} \boldsymbol{\Sigma}_p \phi(\textbf{Z}^\prime)$. 
Let $\textbf{Z}_{1:t}=(\textbf{Z}^{(1)},\ldots,\textbf{Z}^{(t)})$ and $\textbf{y}_t=(y_{\texttt{obs}}^{(1)},\ldots,y_{\texttt{obs}}^{(t)})^{\mbox{\tiny{T}}}$ collect the evaluated inputs and outputs. Substituting the kernel yields the vanilla GP posterior at an unseen $\textbf{Z}_*$:
\begin{equation}
    \begin{split}
        & f_* \mid \textbf{Z}_{1:t}, \textbf{y}_t, \textbf{Z}_* \sim \mathcal{N}(\mu_t(\textbf{Z}_*), \sigma_t^2(\textbf{Z}_*)), \\
        & \text{where } \mu_t(\textbf{Z}_*) = \textbf{k}_t^{\mbox{\tiny{T}}}(\textbf{Z}_*) [\textbf{K}_t + \boldsymbol{\Sigma}_{\epsilon,t}]^{-1} \textbf{y}_t, \\
        & \sigma_t^2(\textbf{Z}_*) = k(\textbf{Z}_*, \textbf{Z}_*) - \textbf{k}_t^{\mbox{\tiny{T}}}(\textbf{Z}_*) [\textbf{K}_t + \boldsymbol{\Sigma}_{\epsilon,t}]^{-1} \textbf{k}_t(\textbf{Z}_*),
    \end{split}
\end{equation}
where $[\textbf{K}_t]_{ij}=k(\textbf{Z}^{(i)},\textbf{Z}^{(j)})$ is the covariance matrix evaluated on $\textbf{Z}_{1:t}$, $\textbf{k}_t(\textbf{Z}_*)$ contains the kernel evaluations between $\textbf{Z}_*$ and all previously evaluated designs, and $\boldsymbol{\Sigma}_{\epsilon,t}=\mathrm{diag}(\sigma_{\epsilon}^2(\textbf{Z}^{(1)}),\ldots,\sigma_{\epsilon}^2(\textbf{Z}^{(t)}))$ is the diagonal observation-noise covariance.
Thus, the GP compares designs through anchored latent coordinates, while admissibility and constraints are still determined by the catalog sections and finite-element model.

\textbf{Acquisition function optimization:} Given the GP posterior, the next step is to decide which admissible catalog combination should be evaluated by MC-FEA.
The exploitation and exploration balance is achieved by designing an acquisition function $\alpha(\textbf{Z}|\mathcal{D}_t)$.
Though numerous acquisition functions exist \cite{shahriari2015taking}, we adopt the Lower Confidence Bound (LCB) acquisition for minimization tasks:
\begin{equation}\label{eq_lcb}
    \alpha_{\texttt{LCB}}(\textbf{Z}|\mathcal{D}_t) = \mu_t(\textbf{Z}) - \kappa \sigma_t(\textbf{Z}),
\end{equation}
where $\kappa\geq0$ is a hyperparameter controlling the exploration level.
In COBALT, this acquisition is never optimized over the surrounding continuous latent space; it is minimized only over anchored catalog configurations that can be assembled and analyzed as finite-element structures.

\subsection{High-dimensional BO with decompositions}
\label{sc2_3}
BO in high-dimensional spaces is an active area of research.
In categorical structural optimization, the dimensionality increases with the number of grouped members or components whose cross-sections must be selected from catalogs.
Decomposition-based strategies are therefore introduced to model a large categorical combination as several lower-dimensional subproblems.
Specifically, a decomposition $g=\{c_1,\ldots,c_{|g|}\}$ over $e$ categorical variables is a collection of non-empty index subsets $c\subseteq\{1,2,\ldots,e\}$.
For any subset $c$, $\textbf{Z}_{[c]}$ denotes the sub-vector of $\textbf{Z}$ containing only the anchored latent coordinates associated with the variables whose indices belong to $c$.
Decomposition-based BO approximates the robust structural response by an additive surrogate,
$f(\textbf{Z}) \approx \sum_{c \in g} f_c\left(\textbf{Z}_{[c]}\right)$.
This approximation is used only for statistical learning; the MC-FEA oracle still evaluates the complete assembled structure.

\textbf{Additive GP kernels:} Compared to standard BO, decomposition methods employ additive kernels \cite{duvenaud2011additive,cheng2019additive} that reflect the above surrogate structure:
$k^g(\textbf{Z}, \textbf{Z}') = \sum_{c \in g} k_c\left(\textbf{Z}_{[c]}, \textbf{Z}'_{[c]}\right)$.
If two categorical variables $i$ and $j$ do not appear together in any of the sets $c$, the kernel does not model their interaction.
For the additive GP, the posterior contribution associated with component $c$ can be written as $p\left(f_c\left(\textbf{Z}_{[c]}\right)\mid \mathcal{D}_t \right) = \mathcal{N}\left(\mu_{t,c}\left(\textbf{Z}_{[c]}\right), \sigma^{2}_{t,c}\left(\textbf{Z}_{[c]}\right) \right)$, where
\begin{equation}\label{eq_lcaab}
\begin{split}
    \mu_{t,c}\left(\textbf{Z}_{[c]}\right) &= \textbf{k}^{\mbox{\tiny{T}}}_{t,c}\left(\textbf{Z}_{[c]}\right) [\textbf{K}_t^g + \boldsymbol{\Sigma}_{\epsilon,t}]^{-1}\textbf{y}_t,\\
    \sigma^{2}_{t,c}\left(\textbf{Z}_{[c]}\right) &= k_{c}\left(\textbf{Z}_{[c]}, \textbf{Z}_{[c]}\right) - \textbf{k}_{t,c}^{\mbox{\tiny{T}}}\left(\textbf{Z}_{[c]}\right)[\textbf{K}_t^g+\boldsymbol{\Sigma}_{\epsilon,t}]^{-1}\textbf{k}_{t,c}\left(\textbf{Z}_{[c]}\right),
\end{split}
\end{equation}
where $\textbf{K}_t^g$ is the additive covariance matrix generated by $k^g$ on the evaluated designs and $\textbf{k}_{t,c}(\textbf{Z}_{[c]})$ is the vector of component-kernel evaluations between $\textbf{Z}_{[c]}$ and the corresponding subvectors of all inputs in $\mathcal{D}_t$.

When each subset $c$ has size strictly less than $e$, the surrogate fits on lower-dimensional subspaces, and the acquisition inherits the additive form $\alpha^{(\text{add-LCB})}_t(\textbf{Z}) = \sum_{c \in g} [\mu_{t,c}(\textbf{Z}_{[c]}) - \kappa \sigma_{t,c}(\textbf{Z}_{[c]})]$. 
This advantage, however, hinges on the choice of $g$.
Existing methods learn $g$ from data via maximum likelihood, but sparse noisy observations from structural OUU may not reliably reveal the true interaction pattern; Section~\ref{sc3} addresses this challenge.

\section{Methodology}
\label{sc3}
\begin{figure}
    \centering
    \includegraphics[width=\textwidth]{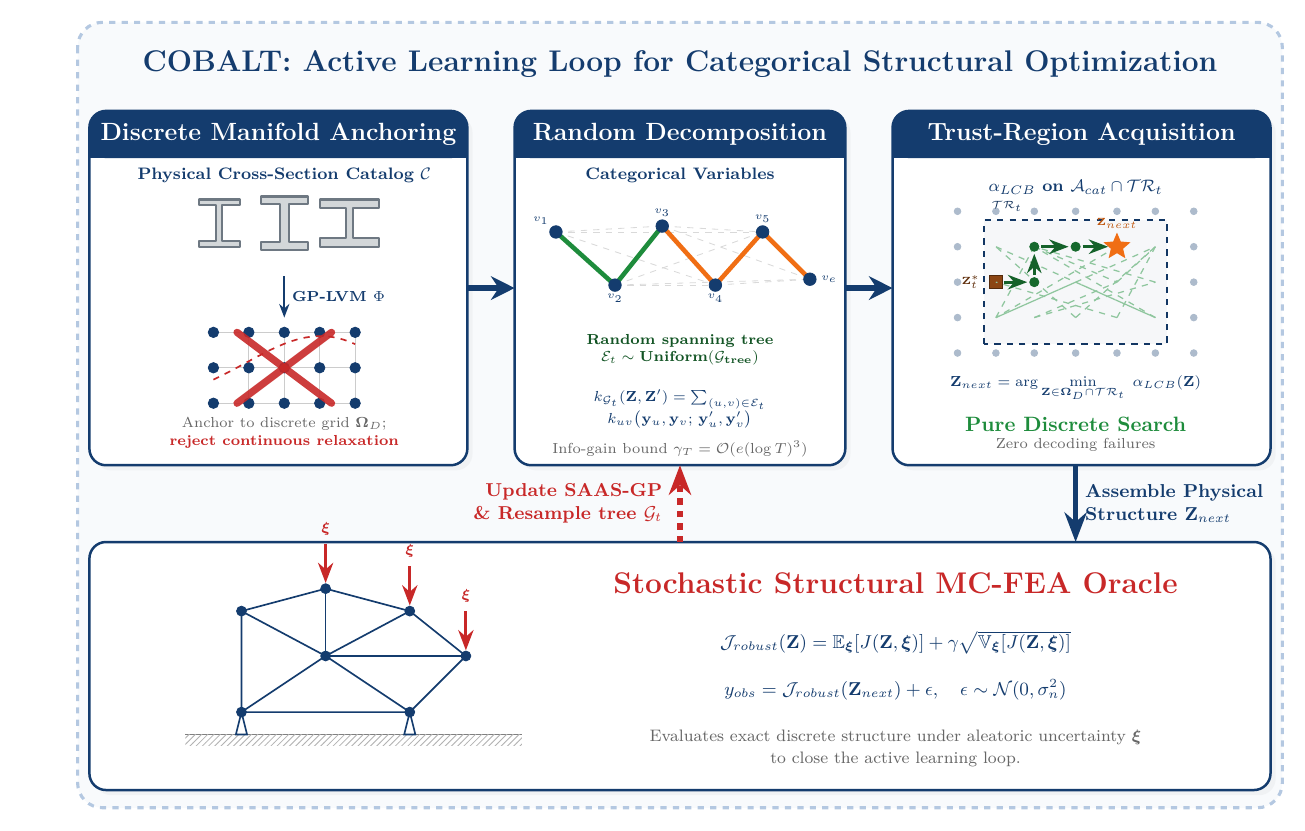}
    \caption{The active learning loop of the proposed COBALT framework.}
    \label{fig:cobalt_framework}
\end{figure}

To mitigate the decoding failures of continuous relaxation and the computational intractability of high-dimensional categorical OUU, we propose the COBALT framework, which treats discreteness as a hard constraint throughout instead of relaxation-and-rounding. 
The methodology comprises four modules, illustrated in Fig.~\ref{fig:cobalt_framework}: First, Isomap maps the nominal section attributes to a low-dimensional latent representation, with the mapped instances locked as a discrete tensor grid $\Omega_D$.
Second, a random tree decomposition models possible interaction patterns among the categorical variables without learning a fragile structure from sparse data.
Third, an additive SAAS-GP surrogate with graph-based evolutionary operators minimizes the LCB acquisition over admissible anchored instances within a dynamically scaled trust region $\mathcal{TR}_t$. 
This step yields the next design $\textbf{Z}_{\texttt{next}}$. 
Finally, the selected configuration is evaluated by MC-FEA under aleatoric uncertainties $\boldsymbol{\xi}$, updating the surrogate and closing the active learning loop.

\subsection{Dimensionality reduction and discrete latent manifold anchoring}
\label{sc3_1}
As defined in Section \ref{sc2}, the available values of any categorical design variable form a finite catalog. For clarity, this subsection first considers a common catalog shared by all variables; variable-specific catalogs follow by replacing $\mathcal{A}_{\texttt{cat}}$ with $\mathcal{A}_{cat,i}$ for each variable $i$.
Each catalog instance is described by a multi-dimensional nominal attribute vector:
\begin{equation}
    \label{Xsapce}
    \begin{split}
        &\textbf{X}=[\textbf{x}^{\mbox{\tiny{1}}},\textbf{x}^{\mbox{\tiny{2}}},\cdots,\textbf{x}^{\mbox{\tiny{{n}}}}],\\
        &\textbf{x}^{\mbox{\tiny{j}}}=({}^{\mbox{\tiny{1}}}a^{\mbox{\tiny{j}}},{}^{\mbox{\tiny{2}}}a^{\mbox{\tiny{j}}},\cdots,{}^{\mbox{\tiny{M}}}a^{\mbox{\tiny{j}}})^{\mbox{\tiny{T}}},\hspace{2mm}\mbox{\normalsize{j}}=1,2,\cdots,{\mbox{\normalsize{{n}}}},
    \end{split}
\end{equation}
where $\textbf{X}\in\mathbb{R}^{M\times n}$ is the nominal catalog matrix and $\textbf{x}^{\mbox{\tiny{j}}}\in\mathbb{R}^{M}$ is the $j$-th available instance. In structural applications, the entries of $\textbf{x}^{\mbox{\tiny{j}}}$ may be the cross-sectional area, moments of inertia, torsional constant, section modulus, or other descriptors required to assemble the element stiffness and capacity checks. To obtain a compact ordering of these discrete instances, we employ non-linear manifold learning (e.g., Isomap). This constructs a deterministic mapping $\Phi: \mathbb{R}^{M} \rightarrow \mathbb{R}^{m}$ ($M>m$), projecting the high-dimensional nominal attributes to a low-dimensional representation:
\begin{equation}
    \label{Ysapce}
    \begin{split}
        &\textbf{X}\xrightarrow{\Phi}\textbf{Y}=[\textbf{y}^{\mbox{\tiny{1}}},\textbf{y}^{\mbox{\tiny{2}}},\cdots,\textbf{y}^{\mbox{\tiny{{n}}}}],\\
        &\textbf{y}^{\mbox{\tiny{j}}}=({}^{\mbox{\tiny{1}}}b^{\mbox{\tiny{j}}},{}^{\mbox{\tiny{2}}}b^{\mbox{\tiny{j}}},\cdots,{}^{\mbox{\tiny{m}}}b^{\mbox{\tiny{j}}})^{\mbox{\tiny{T}}},\hspace{2mm}\mbox{\normalsize{j}}=1,2,\cdots,{\mbox{\normalsize{{n}}}}.
    \end{split}
\end{equation}

To preserve the relative locations of the available instances in the physical catalog, $\Phi$ minimizes the distance-preservation error between the original high-dimensional geodesic distances and the latent distances. The attributes are first normalized component-wise within $[0, 1]$ when their magnitudes differ substantially, so that no single physical descriptor dominates the latent layout.

We mathematically lock the mapped coordinates, defining them strictly as an absolute anchor set in the latent space:
\begin{equation}
    \label{anchor_set}
    \mathcal{A}_{\texttt{cat}} = \{\textbf{y}^{\mbox{\tiny{1}}}, \textbf{y}^{\mbox{\tiny{2}}}, \cdots, \textbf{y}^{\mbox{\tiny{n}}}\} \subset \mathbb{R}^{m}.
\end{equation}

Consequently, for a structural system comprising $e$ categorical variables, a design is formed by choosing one anchored instance for each variable. Let $s_i\in\{1,\ldots,n\}$ be the catalog index selected by variable $i$. The corresponding physical design is $(\textbf{x}^{s_1},\ldots,\textbf{x}^{s_e})$, and its latent representative is
\begin{equation}
    \label{tensor_grid}
    \textbf{Z}(s_1,\ldots,s_e) = [(\textbf{y}^{s_1})^{\mbox{\tiny{T}}},(\textbf{y}^{s_2})^{\mbox{\tiny{T}}},\cdots,(\textbf{y}^{s_e})^{\mbox{\tiny{T}}}]^{\mbox{\tiny{T}}} \in \Omega_D \equiv (\mathcal{A}_{\texttt{cat}})^{e} \subset \mathbb{R}^{D},
\end{equation}
where the total latent search dimensionality is $D = m \times e$. The mapping between $(s_1,\ldots,s_e)$ and $\textbf{Z}\in\Omega_D$ is one-to-one because every coordinate block of $\textbf{Z}$ must be an anchored catalog point. Any coordinate $\textbf{Z} \notin \Omega_D$ is physically inadmissible. Thus, every candidate considered by COBALT corresponds to a combination of available catalog instances before any structural analysis is performed.

\textbf{Remark (Deterministic geometry and stochastic optimization).}
A deterministic catalog mapping may appear counterintuitive within a Bayesian uncertainty-aware framework. 
The tension dissolves once the nominal design description is separated from its aleatoric physical realization. 
The catalog $\textbf{X}$ stores the nominal section attributes that define the admissible choices, while $\boldsymbol{\xi}$ injects multiplicative random perturbations into the MC-FEA oracle. 
These perturbations enter the structural responses during evaluation but leave the fixed anchors $\mathcal{A}_{\texttt{cat}}$ unchanged.

Each source of uncertainty is therefore routed to its appropriate treatment module.
The aleatoric uncertainties in section properties, material properties, loads, and where applicable geometry are propagated through stochastic structural simulation and summarized by $\mathcal{J}_{\texttt{robust}}$. The finite-sample MC-FEA error is represented by the heteroscedastic observation noise $\epsilon(\textbf{Z})$ in Eq.~\eqref{noise_model}. The epistemic uncertainties of the surrogate approximation to $f$ is propagated through the fully Bayesian posterior $p(f_{*}\mid \textbf{Z},\mathcal{D}_{t})$ and expressed via the predictive standard deviation $\sigma_{t}(\textbf{Z})$ in the acquisition function. The lack of reliable prior knowledge about the additive decomposition structure is addressed through the uniform random spanning-tree scheme $S_{\text{r}}$ described below.

Replacing the deterministic Isomap layer with a stochastic alternative, such as a Gaussian Process Latent Variable Model, would introduce posterior variability into the anchor set $\mathcal{A}_{\texttt{cat}}$. This would complicate the definition of both the tensor grid $\Omega_D$ and the trust region $\mathcal{TR}_t$, which require $\Omega_D$ to be fixed. The deterministic mapping is therefore a deliberate modeling choice for routing each uncertainties source to its proper treatment module while preserving a fixed catalog of physically admissible designs.

\subsection{Data-independent random tree decompositions}
\label{sc3_2}
Evaluating the robust objective $\mathcal{J}_{\texttt{robust}}$ over the immense combinatorial catalog $\Omega_D$ (e.g., $n^e$ combinations) necessitates a decomposition-based surrogate model to mitigate the curse of dimensionality. 

\textbf{Misleading decomposition learners:} Conventional decomposition methods typically attempt to learn the optimal additive kernel structure from collected data by maximizing the marginal likelihood. However, relying entirely on local MC-FEA data to infer the interaction topology among categorical variables is highly problematic. In structural OUU, changing one catalog instance may appear to affect only a local member group. However, the same replacement can trigger strongly coupled stress, displacement, or buckling responses in distant parts of the structure. Consequently, a data-driven learner can exploit this local view, falsely deduce a complete separation among variables, and become trapped in suboptimal local modes.

\textbf{Adversarial formulation and data-independent rules:} To bypass this misleading phenomenon, we formulate the decomposition from an adversarial perspective. Let $\mathfrak{G}$ denote a prescribed class of admissible decomposition graphs over the variable set $\mathcal{V}=\{1,\ldots,e\}$. Each graph $G=(\mathcal{V},\mathcal{E})\in\mathfrak{G}$ defines an additive RKHS $\mathcal{H}^{G}$ through the edge-wise kernel components. Instead of learning $G$ from the evaluated catalog combinations, we introduce a predefined, data-independent randomized scheme $S_{\text{r}}$. This scheme samples a graph $\mathcal{G}_{t}\in\mathfrak{G}$ at each active learning round $t$ according to a prescribed distribution.

The role of the decomposition can be interpreted through the quantities that appear in standard upper-confidence-bound-style analyses. The first is the maximum information gain
$\Gamma_T=\max_{\mathcal{Q}\subset\Omega_D,\,|\mathcal{Q}|=T} I(\textbf{f}_{\mathcal{Q}};\textbf{y}_{\mathcal{Q}})$, where $\textbf{f}_{\mathcal{Q}}$ and $\textbf{y}_{\mathcal{Q}}$ denote the latent true responses and noisy MC-FEA observations over a finite queried set $\mathcal{Q}$. This term measures the statistical complexity of the selected additive kernel. The second is the approximation mismatch $\varepsilon_t = \sup_{\textbf{Z} \in \Omega_D} |\hat{f}_t(\textbf{Z}) - f(\textbf{Z})|$, where $\hat{f}_t \in \mathcal{H}^{\mathcal{G}_t}$ is the closest function to the true robust response $f$ within the RKHS induced by the graph selected at iteration $t$. Under the regularity assumptions used in additive-kernel GP-UCB analyses, the following high-probability scaling motivates the decomposition design:
\begin{equation}
    \label{regret_bound}
    R_T = \mathcal{O} \left( \sqrt{T\Gamma_T} \left( B + \sqrt{\ln\frac{1}{\delta_A} + \Gamma_T} + \frac{\mathbb{E}_{S_{\text{r}}} \left[\sum_{t=1}^T \varepsilon_t \right]}{\delta_B} \right) \right),
\end{equation}
where $R_T$ denotes the cumulative regret over $T$ active learning rounds, $I(\cdot;\cdot)$ denotes mutual information, $B = \max_{t \le T} \| \hat{f}_t \|_{\mathcal{H}^{\mathcal{G}_t}}$ is an RKHS norm bound, and $\delta_A, \delta_B \in (0,1)$ are high-probability confidence parameters.
This expression is used here as a design guideline rather than a complete convergence theorem for the implemented constrained optimizer. It indicates that an effective decomposition scheme should simultaneously restrict the information gain $\Gamma_T$ and reduce the expected mismatch $\mathbb{E}_{S_{\text{r}}}[\sum_{t=1}^T \varepsilon_t]$, where $\mathbb{E}_{S_{\text{r}}}[\cdot]$ denotes expectation with respect to the randomized graph-selection scheme.

\textbf{Analysis of decomposition rules:} To control $\Gamma_T$, we restrict $\mathfrak{G}$ to graphs with exclusively pairwise components. If $\mathfrak{G}$ is further restricted to tree-based decompositions spanning the $e$ categorical variables, the maximum information-gain scaling can be bounded as $\Gamma_T \le \mathcal{O}(e(\log T)^3)$ under standard assumptions for additive kernels. This is lower than the cost of incorporating fully connected pairwise interactions.

While tree-based topologies bound the kernel complexity, reducing the expected mismatch $\mathbb{E}_{S_{\text{r}}}[\sum_{t=1}^T \varepsilon_t]$ against an unknown structural black-box remains challenging. If $S_{\text{r}}$ degenerates to the same fixed tree decomposition at every iteration, a highly coupled structural response may repeatedly appear through omitted interactions, resulting in persistent mismatch. Conversely, adaptively increasing the interaction complexity destroys the tree property and inflates $\Gamma_T$. To balance these two effects without assuming a reliable prior interaction graph, COBALT uses a neutral randomized rule:
\begin{equation}
    \label{randomized_scheme}
    \Pr(\mathcal{G}_t = G) = \frac{1}{|\mathfrak{G}|}, \quad G \in \mathfrak{G}.
\end{equation}

This randomized tree-selection principle does not claim that one tree is universally optimal for a known objective. Instead, it avoids committing the surrogate to a single potentially misleading decomposition when the available MC-FEA data are sparse, noisy, and locally biased. By uniformly sampling a random tree at each iteration, COBALT repeatedly changes the pairwise view through which catalog combinations are modeled, while keeping the complexity bounded in high-dimensional discrete spaces.

\subsection{uncertainties-aware additive sparse surrogate modeling (Additive-SAAS-GP)}
\label{sc3_3}
At optimization iteration $t$, the evaluated data consist only of admissible categorical configurations and their MC-FEA response estimators, denoted by $\mathcal{D}_t = \{(\textbf{Z}^{(i)}, y_{\texttt{obs}}^{(i)})\}_{i=1}^t$ with $\textbf{Z}^{(i)}\in\Omega_D$. Due to the stochastic nature of uncertainties propagation, the observations yield heteroscedastic noise governed by the Monte Carlo sampling variance:
\begin{equation}
    \label{noise_model}
    y_{\texttt{obs}}^{(i)} = f(\textbf{Z}^{(i)}) + \epsilon^{(i)}, \quad \epsilon^{(i)} \sim \mathcal{N}(0, \sigma_{\epsilon}^2(\textbf{Z}^{(i)})),
\end{equation}
where $f(\textbf{Z})$ represents the underlying true robust objective. 

Building upon the data-independent randomized scheme $S_{\text{r}}$, we construct a fully Bayesian additive surrogate $f(\textbf{Z}) \sim \mathcal{GP}(\mu_0, k_{\mathcal{G}_t}(\textbf{Z}, \textbf{Z}'))$ with a zero prior mean $\mu_0 = 0$. Given the uniformly sampled random tree $\mathcal{G}_t = (\mathcal{V}, \mathcal{E}_t)$ with $|\mathcal{V}| = e$ nodes and $|\mathcal{E}_t| = e-1$ edges, the similarity between two catalog designs is measured by comparing the anchored latent coordinates of their selected instances. Let $\textbf{Z}=[(\textbf{y}_{\mbox{\tiny{1}}})^{\mbox{\tiny{T}}},\ldots,(\textbf{y}_{\mbox{\tiny{e}}})^{\mbox{\tiny{T}}}]^{\mbox{\tiny{T}}}$ and $\textbf{Z}'=[(\textbf{y}'_{\mbox{\tiny{1}}})^{\mbox{\tiny{T}}},\ldots,(\textbf{y}'_{\mbox{\tiny{e}}})^{\mbox{\tiny{T}}}]^{\mbox{\tiny{T}}}$ denote two anchored designs, where $\textbf{y}_{\mbox{\tiny{i}}},\textbf{y}'_{\mbox{\tiny{i}}}\in\mathcal{A}_{\texttt{cat}}$ are the section coordinates selected by variable $i$. This leads to an additive kernel decomposed strictly over the tree edges:
\begin{equation}
    \label{additive_kernel}
    k_{\mathcal{G}_t}(\textbf{Z}, \textbf{Z}') = \sum_{(u,v) \in \mathcal{E}_t} k_{uv}\left( [(\textbf{y}_{\mbox{\tiny{u}}})^{\mbox{\tiny{T}}}, (\textbf{y}_{\mbox{\tiny{v}}})^{\mbox{\tiny{T}}}]^{\mbox{\tiny{T}}}, [(\textbf{y}'_{\mbox{\tiny{u}}})^{\mbox{\tiny{T}}}, (\textbf{y}'_{\mbox{\tiny{v}}})^{\mbox{\tiny{T}}}]^{\mbox{\tiny{T}}} \right),
\end{equation}

Each base kernel $k_{uv}$ is an Automatic Relevance Determination (ARD) kernel operating exclusively on the localized $2m$-dimensional latent subspace of the two connected categorical variables. 

To induce feature sparsity within the active pairwise subspaces, a SAAS prior is imposed on the inverse lengthscales (feature weights $\theta_d = \rho_d^{-1}$) of the base kernels. It is implemented through a heavy-tailed hierarchical Horseshoe distribution:
\begin{equation}
    \label{horseshoe}
    \theta_d \sim \mathcal{HC}(0, \tau), \quad \tau \sim \mathcal{HC}(0, \tau_0),
\end{equation}
where $\mathcal{HC}$ denotes the half-Cauchy distribution, $\rho_d$ is the lengthscale of the $d$-th latent coordinate in the active subspace, $\tau$ is the global shrinkage parameter, and $\tau_0$ is its scale hyperparameter.
Under fully Bayesian inference via the No-U-Turn Sampler, the hyperparameter posteriors $\boldsymbol{\Theta} = \{\sigma_f^2, \boldsymbol{\theta}, \boldsymbol{\sigma}_{\epsilon,t}^2, \tau\}$ are marginalized rather than point-estimated. Here $\sigma_f^2$ is the kernel output variance, $\boldsymbol{\theta}$ collects the inverse lengthscales, and $\boldsymbol{\sigma}_{\epsilon,t}^2$ collects the observation-noise variances. The predictive posterior distribution for any unseen but admissible categorical configuration $\textbf{Z}\in\Omega_D$ is integrated over the Markov Chain Monte Carlo (MCMC) hyperparameter samples:
\begin{equation}
    \label{mcmc_integration}
    p(f_* \mid \textbf{Z}, \mathcal{D}_t) = \int p(f_* \mid \textbf{Z}, \mathcal{D}_t, \boldsymbol{\Theta}) p(\boldsymbol{\Theta} \mid \mathcal{D}_t) d\boldsymbol{\Theta},
\end{equation}
where $f_*$ denotes the latent robust objective value at $\textbf{Z}$. This additive hierarchical structure shrinks the weights of irrelevant latent features while decomposing the predictive mean $\mu_t(\textbf{Z})$ and epistemic variance $\sigma_t^2(\textbf{Z})$ into pairwise contributions over the sampled tree. Consequently, the surrogate focuses on influential catalog attributes and structural variable interactions while remaining stable under heteroscedastic MC-FEA noise.

\subsection{Trust-region discrete graph acquisition minimization}
\label{sc3_4}
Based on the decomposed predictive moments $\mu_t$ and $\sigma_t$ extracted from Eq. \eqref{mcmc_integration}, we construct the LCB acquisition function to decide which admissible catalog combination should be evaluated next. Inheriting the pairwise factorization over the sampled random tree edge set $\mathcal{E}_t$, the acquisition over the discrete catalog domain is given by:
\begin{equation}
    \label{lcb_acq}
    \alpha_{\texttt{LCB}}(\textbf{Z}) = \sum_{(u,v) \in \mathcal{E}_t} \left[ \mu_{uv}(\textbf{y}_{\mbox{\tiny{u}}}, \textbf{y}_{\mbox{\tiny{v}}}) - \kappa \cdot \sigma_{uv}(\textbf{y}_{\mbox{\tiny{u}}}, \textbf{y}_{\mbox{\tiny{v}}}) \right],
\end{equation}
where $\mu_{uv}$ and $\sigma_{uv}$ are the posterior mean and standard-deviation contributions of the edge component $(u,v)$, and $\kappa\geq0$ regulates the exploration trade-off. 

To avoid evaluating too many remote combinations in the vast $n^e$ discrete catalog space, COBALT incorporates a dynamically scaled hypercubic trust region ($\mathcal{TR}_t$) centered at the incumbent robust best feasible design $\textbf{Z}^*_{t-1}$. The trust region limits the admissible neighboring catalog combinations considered around the current best design. It is mathematically defined in the normalized latent coordinates via the Chebyshev distance ($L_\infty$-norm):
\begin{equation}
    \label{trust_region}
    \mathcal{TR}_t = \left\{ \textbf{Z} \in \mathbb{R}^D \;\middle|\; \|\textbf{Z} - \textbf{Z}^*_{t-1}\|_{\infty} \le \frac{L_t}{2} \right\},
\end{equation}
where the trust region length $L_t$ expands after a sequence of successful improvements and contracts following consecutive evaluation failures.

Traditional methods often use continuous optimizers followed by nearest-neighbor rounding off.
In contrast, COBALT identifies the next configuration $\textbf{Z}_{\texttt{next}}$ through a purely discrete combinatorial optimization problem over the valid intersection set $\Omega_{TR}^{(t)}$:
\begin{equation}
    \label{discrete_opt}
    \textbf{Z}_{\texttt{next}} = \underset{\textbf{Z} \in \Omega_{TR}^{(t)}}{\mathrm{arg\,min}}\hspace{2mm} \alpha_{\texttt{LCB}}(\textbf{Z}), \quad \text{where } \Omega_{TR}^{(t)} = \Omega_D \cap \mathcal{TR}_t.
\end{equation}

To efficiently solve Eq. \eqref{discrete_opt} over the discrete combinatorial intersection $\Omega_{TR}^{(t)}$ without resorting to any continuous relaxation, we repurpose graph-based evolutionary operators on the anchored catalog network as the internal acquisition search engine. 

Evaluating the additively decomposed algebraic formulation $\alpha_{\texttt{LCB}}(\textbf{Z})$ is inexpensive compared with an expensive MC-FEA in the tested cases. Therefore, the evolutionary operators can explore many alternative combinations of neighboring catalog instances as an internal surrogate search. The crossover and mutation trajectories are constrained to the geodesic pathways of the discrete anchored network $\mathcal{A}_{\texttt{cat}}$ and the dynamic boundary $\mathcal{TR}_t$. This constraint ensures that the identified minimizer $\textbf{Z}_{\texttt{next}}$ is still composed of physically admissible categorical instances. 

This strictly valid configuration $\textbf{Z}_{\texttt{next}}$ is then physically evaluated by MC-FEA without any artificial rounding-off.
This evaluation closes the robust Bayesian active learning loop.
It also avoids the inadmissible intermediate designs that appear in continuous-relaxation strategies.

\subsection{Categorical optimization processes}
\label{sc3_5}
The preceding modules are assembled into a sequential categorical optimization process in Algorithm~\ref{alg:cobalt}, comprising one deterministic catalog-preparation stage and one uncertainty-aware Bayesian search stage. In the preparation stage, the nominal physical catalog is normalized, mapped, and locked as the discrete anchor set $\mathcal{A}_{\texttt{cat}}$, which defines the immutable feasible tensor grid $\Omega_D$. This anchoring establishes the central invariant of COBALT: every subsequent proposal must be an admissible combination of catalog instances. At each active learning iteration, COBALT then samples a data-independent random tree to fix the additive kernel structure. The Additive SAAS-GP surrogate is fitted to the accumulated MC-FEA observations, and the LCB acquisition is built on the sampled tree. The trust region restricts the candidate set to $\Omega_{TR}^{(t)} = \Omega_D \cap \mathcal{TR}_t$, so LCB minimization runs only over physically valid anchored configurations. The selected catalog combination is evaluated by the MC-FEA oracle for its noisy robust objective and feasibility margins. The outcome updates the incumbent feasible optimum and the trust-region length.

\begin{algorithm}[h]
 \caption{COBALT for categorical structural optimization under high-dimensional uncertainties}
   \label{alg:cobalt}
    \begin{algorithmic}[1]
        \State {\bfseries Input:} Categorical catalog $\textbf{X}$ ($n$ instances), variables $e$, evaluations $T$, parameter $\kappa$
        \State {\bfseries Output:} Robust best feasible categorical design $\textbf{Z}^*_T$
        
        \State {\bfseries // Discrete Manifold Anchoring}
        \State Normalize attributes of $\textbf{X}$ column-wise within $[0, 1]$
        \State Apply Isomap $\Phi$ mapping $\textbf{X}$ to latent representation $\textbf{Y}$
        \State Lock mapped coordinates as discrete anchor set $\mathcal{A}_{\texttt{cat}} = \{\textbf{y}^1, \dots, \textbf{y}^n\} \subset \mathbb{R}^m$
        \State Define feasible search domain as discrete tensor grid $\Omega_D \equiv (\mathcal{A}_{\texttt{cat}})^e$
        \State Evaluate initial random designs via MC-FEA to initialize dataset $\mathcal{D}_0$
        \State Identify incumbent best feasible design $\textbf{Z}^*_0$, initialize trust-region length $L_1$
        
        \For{$t = 1, \dots, T$}
            \State {\bfseries // Random Decomposition}
            \State Sample uniform random spanning tree graph $\mathcal{G}_t = (\mathcal{V}, \mathcal{E}_t)$
            
            \State {\bfseries // Trust-Region Discrete Graph Acquisition}
            \State Train Additive SAAS-GP surrogate on $\mathcal{D}_{t-1}$ using tree-decomposed kernel $k_{\mathcal{G}_t}$
            \State Marginalize hyperparameters via MCMC for predictive mean $\mu_t(\textbf{Z})$ and variance $\sigma_t^2(\textbf{Z})$
            \State Construct additive LCB acquisition function factorized over $\mathcal{E}_t$: 
            \State \quad $\alpha_{\texttt{LCB}}(\textbf{Z}) = \sum_{(u,v) \in \mathcal{E}_t} \left[ \mu_{uv}(\textbf{y}_u, \textbf{y}_v) - \kappa \cdot \sigma_{uv}(\textbf{y}_u, \textbf{y}_v) \right]$
            \State Define dynamically scaled hypercubic trust region $\mathcal{TR}_t$ centered at $\textbf{Z}^*_{t-1}$
            \State Define constrained search space intersection $\Omega_{TR}^{(t)} = \Omega_D \cap \mathcal{TR}_t$
            \State {\bfseries Discrete Search:} Use evolutionary operators traversing $\mathcal{A}_{\texttt{cat}}$ to minimize $\alpha_{\texttt{LCB}}$:
            \State \quad $\textbf{Z}_{\texttt{next}} = \underset{\textbf{Z} \in \Omega_{TR}^{(t)}}{\arg\min} \ \alpha_{\texttt{LCB}}(\textbf{Z})$
            
            \State {\bfseries // Stochastic MC-FEA Oracle}
            \State Physically evaluate exact discrete structure $\textbf{Z}_{\texttt{next}}$ via MC-FEA under uncertainties $\boldsymbol{\xi}$
            \State Obtain heteroscedastic noisy robust observation $y_{\texttt{obs}}^{(t)} = f(\textbf{Z}_{\texttt{next}}) + \epsilon^{(t)}$ and feasibility margins $\textbf{h}(\textbf{Z}_{\texttt{next}})$
            \State Augment dataset: $\mathcal{D}_t = \mathcal{D}_{t-1} \cup \{(\textbf{Z}_{\texttt{next}}, y_{\texttt{obs}}^{(t)})\}$
            \State Update robust incumbent best feasible design $\textbf{Z}^*_t$ subject to $\textbf{h}(\textbf{Z}) \leq \textbf{0}$
            \State Expand or contract trust-region length $L_{t+1}$ based on recent performance
        \EndFor
        \State {\bfseries Return} Optimal robust structurally feasible design $\textbf{Z}^*_T$
    \end{algorithmic}
\end{algorithm}

Algorithm~\ref{alg:cobalt} highlights three operational safeguards that distinguish COBALT from continuous latent-space categorical optimization. First, the anchor set $\mathcal{A}_{\texttt{cat}}$ and tensor grid $\Omega_D$ remain fixed after the initial mapping, so the available instances and their adjacency relations are never changed during optimization. Second, the decomposition graph $\mathcal{G}_t$ is resampled independently of the observed objective values, avoiding the instability of likelihood-driven structure learners under sparse and noisy MC-FEA data. Third, the acquisition search is constrained to traverse anchored graph paths inside $\Omega_{TR}^{(t)}$. As a result, the recommended point $\textbf{Z}_{\texttt{next}}$ is already a physically decodable design before the expensive oracle is called.

\section{Numerical examples}
\label{sc4}
Five numerical examples validate COBALT, as shown in Tab.~\ref{tab:information}. The uncertain vector $\boldsymbol{\xi}$ aggregates section-attribute, material, and load perturbations, plus geometric imperfections. All examples share the same evaluation budget and Monte Carlo sample size. The catalog, uncertainties model, MC-FEA sample size, robust objective definition, feasibility checks, and total number of expensive structural evaluations are kept identical across the compared methods. For Bayesian-optimization-based methods, the same surrogate family and hyperprior settings are used. The key difference is the acquisition-search domain: COBALT searches over anchored catalog configurations, whereas the continuous latent-space baseline optimizes in the surrounding continuum and then applies nearest-neighbor decoding. The convergence histories report the best feasible incumbent found as a function of expensive MC-FEA calls; internal acquisition evaluations are not counted as structural evaluations.
\begin{table}[t]
    \centering
    \caption{The information of the five numerical examples.}
    \label{tab:information}
    \begin{tabular}{cccccc}
        \toprule
        Name & No. of beams & No. of design variables & Problem type & Attributes & Description \\
        \midrule
        10-beam     & 10   & 4 & Planar & $A$, $I_{\mbox{\tiny{y}}}$, $I_{\mbox{\tiny{z}}}$& Validity benchmark test\\
        Dome        & 120  & 7 & Spatial & $A$, $I_{\mbox{\tiny{y}}}$, $I_{\mbox{\tiny{z}}}$, $J_{\mbox{\tiny{x}}}$ &  3D benchmark test\\
        % Six-story   & 63   & 8 & Spatial & $A$, $I_{\mbox{\tiny{y}}}$, $I_{\mbox{\tiny{z}}}$, $J_{\mbox{\tiny{x}}}$ & 3D benchmark test   \\
        105-beam    & 105  & 105 & Planar & $A$, $I_{\mbox{\tiny{y}}}$, $I_{\mbox{\tiny{z}}}$ & Medium-scale variable test   \\
        1564-beam   & 1564 & 1564 & Spatial &$A$, $I_{\mbox{\tiny{y}}}$, $I_{\mbox{\tiny{z}}}$, $J_{\mbox{\tiny{x}}}$ & Large-scale variable test   \\
        \bottomrule
    \end{tabular}
\end{table}

The COBALT framework is benchmarked against the following baseline methods:
\begin{itemize}
\item \textbf{Continuous latent-space Bayesian optimization:} A standard Latent Space BO with polynomial manifold interpolation, gradient-based acquisition optimization, and nearest-neighbor rounding to the discrete catalog.
\item \textbf{Deterministic manifold optimization:} A deterministic search over the mapped catalog manifold, without MC-FEA-driven Bayesian active learning, that decodes the latent solution to a catalog instance.
\end{itemize}

For all Bayesian-optimization-based methods, the SAAS-GP surrogate of Section~\ref{sc3_3} is used with identical hyperprior settings for a fair comparison. The key differentiator is the acquisition strategy: COBALT traverses only the discrete anchored graph, whereas continuous latent-space Bayesian optimization optimizes in the latent continuum and then rounds off. Comparisons are read off the best feasible robust objective, constraint satisfaction, and convergence within the prescribed budgets.

\subsection{The planar ten-beam truss}
\label{sc4_1}
\begin{figure}
    \centering
    \includegraphics[width=0.45\columnwidth]{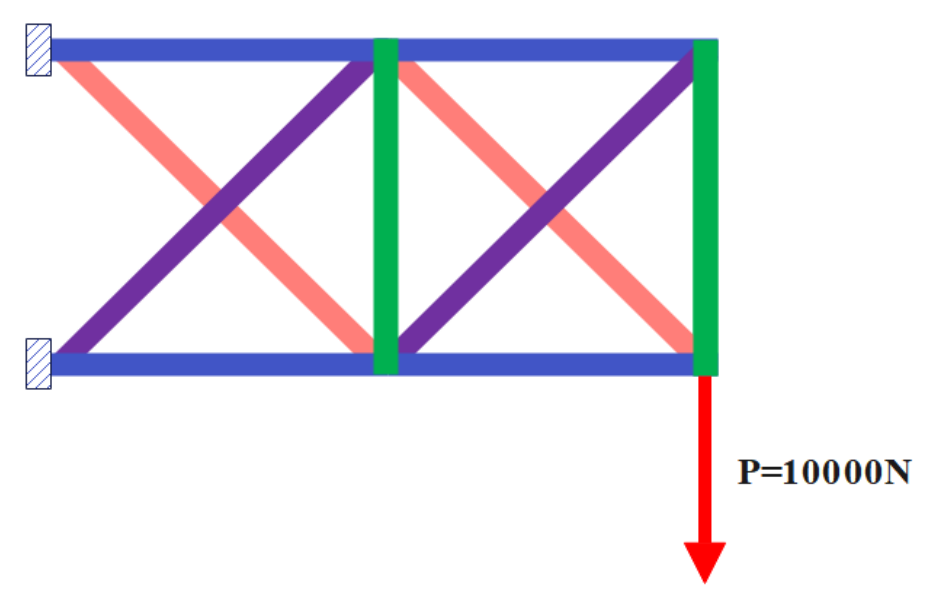}
    \caption{The load-case illustration of the ten-beam structure.} 
    \label{fig:tenbar}
\end{figure}
The first benchmark is the classical ten-beam planar truss. The truss consists of 10 beam members connected at 6 nodes, as illustrated in Fig. \ref{fig:tenbar}. The 10 beams are divided into 4 groups: the horizontal group, the vertical group, the sub-diagonal group, and the principal diagonal group. Each group shares the same cross-section. The left two nodes are fixed on a rigid wall and the lower-right node bears a downward load of 10000\,N. Each group is assigned a categorical cross-section variable $\textbf{x}_i$ ($i = 1, \ldots, 4$) selected from a predefined catalog of 49 standard steel profiles. The catalog instances differ simultaneously in cross-sectional area $A$ and the moments of inertia $I_{\mbox{\tiny{y}}}$ and $I_{\mbox{\tiny{z}}}$, forming a multi-dimensional attribute representation. The material density is $\rho = 7850\,\mathrm{kg/m^3}$ and the gravitational acceleration is $g = 9.81\,\mathrm{m/s^2}$.
The robust optimization problem is formulated as:
\begin{equation}
    \label{tenbar_obj}
    \begin{split}
&\text{min.:}\hspace{5mm}\mathcal{J}_{\texttt{robust}}(\textbf{x}_{\mbox{\tiny{1}}},\textbf{x}_{\mbox{\tiny{2}}},\textbf{x}_{\mbox{\tiny{3}}},\textbf{x}_{\mbox{\tiny{4}}})=\mathbb{E}_{\boldsymbol{\xi}}\left[\frac{1}{2}\textbf{u}^{\mbox{\tiny{T}}}\textbf{K}\textbf{u}\right] + \gamma \sqrt{\mathbb{V}_{\boldsymbol{\xi}}\left[\frac{1}{2}\textbf{u}^{\mbox{\tiny{T}}}\textbf{K}\textbf{u}\right]};\\
&\text{s.}\hspace{1.5mm}\text{t.:}\hspace{5.8mm}mass(\textbf{x}_{\mbox{\tiny{1}}},\textbf{x}_{\mbox{\tiny{2}}},\textbf{x}_{\mbox{\tiny{3}}},\textbf{x}_{\mbox{\tiny{4}}})-240\leq 0;\\
&\hspace{13mm}\textbf{K}\textbf{u}=\textbf{P};\\
&\hspace{13mm}\max(\textbf{F}-\textbf{F}_y^{\mbox{\tiny{cr}}})\leq 0,\hspace{2mm}\textbf{F}_y^{\mbox{\tiny{cr}}}=(f_y^{\mbox{\tiny{cr1}}},f_y^{\mbox{\tiny{cr2}}},\cdots,f_y^{\mbox{\tiny{cr10}}});\\
&\hspace{13mm}\max(\textbf{F}-\textbf{F}_z^{\mbox{\tiny{cr}}})\leq 0,\hspace{2mm}\textbf{F}_z^{\mbox{\tiny{cr}}}=(f_z^{\mbox{\tiny{cr1}}},f_z^{\mbox{\tiny{cr2}}},\cdots,f_z^{\mbox{\tiny{cr10}}});\\
&\hspace{13mm}\textbf{x}_{\mbox{\tiny{i}}}\in \{\textbf{x}^{\mbox{\tiny{1}}}_{\mbox{\tiny{i}}},\textbf{x}^{\mbox{\tiny{2}}}_{\mbox{\tiny{i}}},\cdots,\textbf{x}^{\mbox{\tiny{49}}}_{\mbox{\tiny{i}}}\},\hspace{2mm}\mbox{\normalsize{i}}=1,2,3,4;\\
&\hspace{13mm}\textbf{x}^{\mbox{\tiny{j}}}_{\mbox{\tiny{i}}}=(A^{\mbox{\tiny{j}}}_{\mbox{\tiny{i}}},I_{y}{}^{\mbox{\tiny{j}}}_{\mbox{\tiny{i}}},I_{z}{}^{\mbox{\tiny{j}}}_{\mbox{\tiny{i}}})^{\mbox{\tiny{T}}},\hspace{2mm}\mbox{\normalsize{j}}=1,2,\cdots,49.\\
\end{split}
\end{equation}
where $\mathcal{J}_{\texttt{robust}}$ is the robust objective combining the expectation and standard deviation of the structural strain energy. $\gamma$ is the weight factor on the standard deviation, set to 1 in this study. For each realization of $\boldsymbol{\xi}$, $\textbf{K}$, $\textbf{P}$, and $\textbf{u}$ denote the assembled stiffness matrix, applied load vector, and displacement vector satisfying the finite-element equilibrium equation $\textbf{K}\textbf{u}=\textbf{P}$. The corresponding strain energy is $\frac{1}{2}\textbf{u}^{\mbox{\tiny{T}}}\textbf{K}\textbf{u}$. $\textbf{x}_{\mbox{\tiny{i}}}$ denotes the $i$-th design variable. $\textbf{F}$ is the member axial-force vector. The vectors $\textbf{F}_y^{\mbox{\tiny{cr}}}$ and $\textbf{F}_z^{\mbox{\tiny{cr}}}$ are the critical buckling loads in the $y$- and $z$-directions, respectively. The max operator is applied to the componentwise buckling-load margin.

The uncertain parameter vector $\boldsymbol{\xi}$ includes three independent multiplicative uncertainties sources:
\begin{itemize}
    \item Section attributes $(A,I_y,I_z)$ of each selected profile, $\mathrm{CoV}=0.05$.
    \item Young's modulus: $\widetilde{E}=E_0\eta_E$, $\mathrm{CoV}(\eta_E)=0.05$ (manufacturing tolerances), with $E_0=2.1\times10^{11}\,\mathrm{Pa}$.
    \item External load: $\widetilde{P}=P_0\eta_P$, $\mathrm{CoV}(\eta_P)=0.01$ (environmental fluctuations), with $P_0=10^{4}\,\mathrm{N}$.
\end{itemize}

\begin{figure}
    \centering
    \includegraphics[width=\columnwidth]{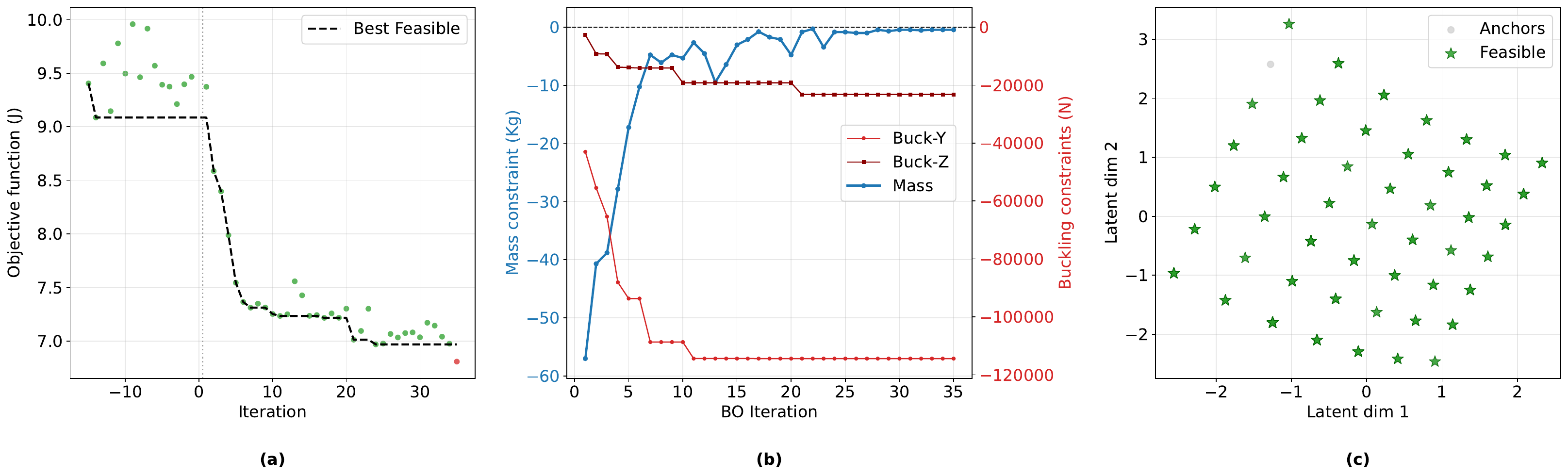}
    \caption{Convergence of the 10-beam optimization: (a) robust objective with the best feasible trajectory, (b) mass and buckling constraints, and (c) feasible anchored designs in the 2D latent space.} 
    \label{fig:tenbar_convergence}
\end{figure}

MC-FEA with $N_{MC}=500$ samples gives the robust metrics for each candidate under heteroscedastic noise (Eq.~\eqref{eq_noise}). Isomap embeds each $M$-dimensional attribute vector into $m$ latent coordinates with $m\ll M$. The anchor set $\mathcal{A}_{\texttt{cat}}$ (Eq.~\eqref{anchor_set}) is the COBALT search space. The full latent dimension is $D=4m$.
Fig. \ref{fig:tenbar_convergence} presents the convergence histories of the robust objective $\mathcal{J}_{\texttt{robust}}$, the buckling constraints in y and z directions and the mass constraint for all iterations over 200 MC-FEA evaluations. After 40 iterations, the robust objective converges and all the design constraints are satisfied.
In this benchmark, COBALT attains the best feasible robust objective within the prescribed budget. 
Continuous latent-space Bayesian optimization keeps pace early on but stagnates due to repeated decoding failures. 
Its continuous acquisition often locates latent pseudo-optima that, after nearest-neighbor rounding, map to structurally inferior or infeasible configurations. 
By contrast, COBALT's discrete graph acquisition guarantees physically admissible recommendations, sustaining steadier improvement throughout the campaign.

% --- The following paragraph (a dedicated discussion of
% Fig.~\ref{fig:tenbar_compre} / 10bar_failure_manifold.pdf) is
% commented out together with the corresponding figure block below
% to fit within the 30-page Engineering Structures limit. The
% disclosure of this removal is recorded in cover-letter.tex.
% Fig. \ref{fig:tenbar_compre} compares the latent-space distributions obtained by different manifold learning methods for the ten-beam optimization problem. The upper-row plots show the mapped catalog instances colored by cross-sectional area, while the lower-row plots report the corresponding failure probability and constraint-violation tendency over the same latent coordinates. The comparison indicates that the quality of the latent representation strongly affects the physical consistency of the categorical search space. Methods that better preserve the topology of the original attribute space produce smoother area gradients and more coherent feasible regions. By contrast, distorted mappings scatter physically dissimilar sections into neighboring latent positions and lead to irregular failure-probability patterns. These results confirm that an appropriate manifold representation is essential for constructing reliable anchored latent graphs and reducing decoding-related constraint violations.

\begin{figure}
    \centering
    \includegraphics[width=\columnwidth]{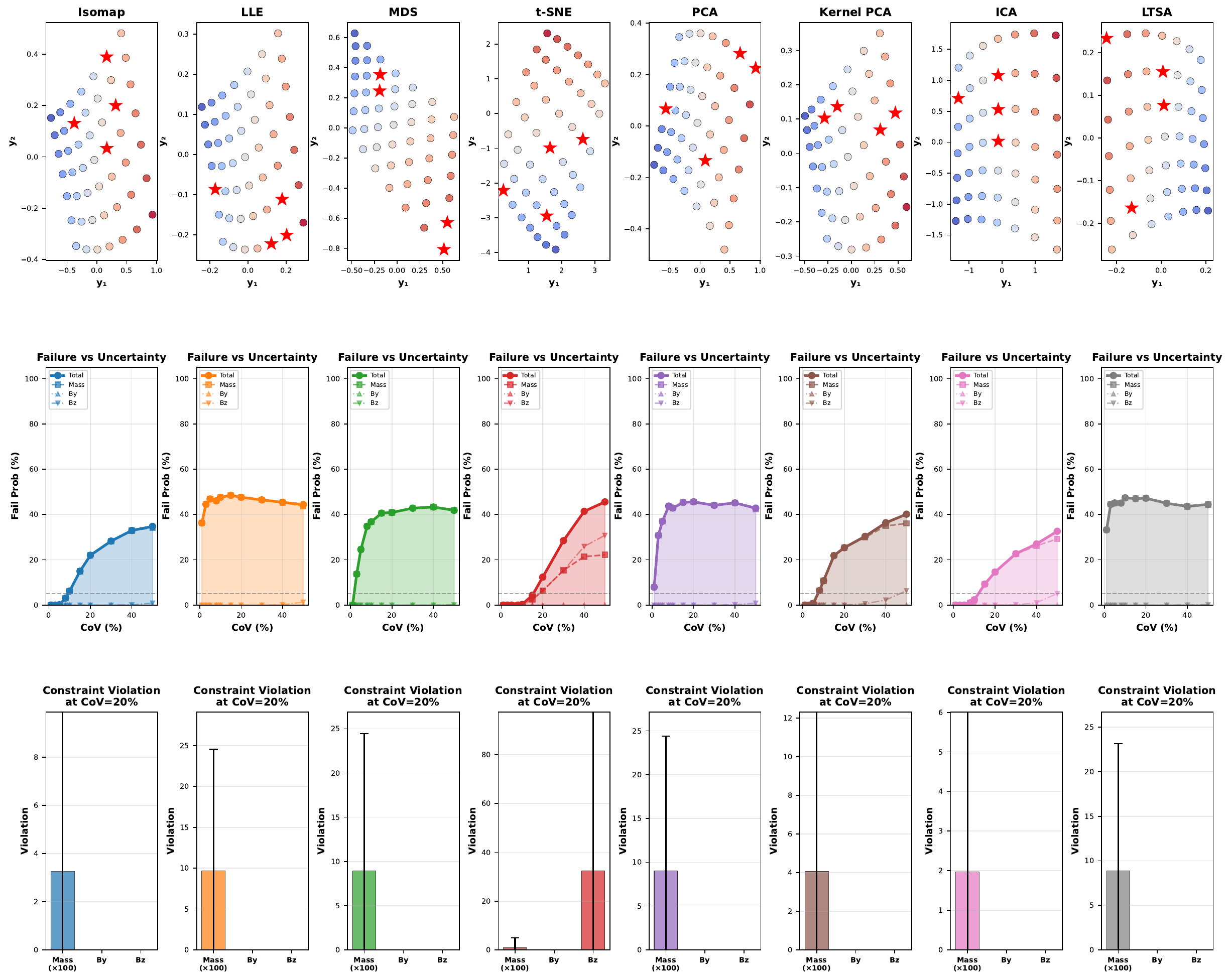}
    \caption{uncertainties sensitivity of deterministic manifold-based optimization for the ten-beam problem.} 
    \label{fig:tenbar_variations}
\end{figure}

Fig. \ref{fig:tenbar_variations} further shows that deterministic manifold-optimization strategies become fragile when uncertainties is introduced. UMAP is not included in this comprehensive uncertainties-sensitivity comparison because of its weaker physical consistency on the present benchmark. As the coefficient of variation increases, the failure probability rises and constraint violations remain method-dependent, indicating that deterministic latent representations alone cannot reliably handle categorical optimization under uncertainties.
% --- Figure 10bar_failure_manifold.pdf (label fig:tenbar_compre) is
% commented out to fit within the 30-page Engineering Structures
% limit. The figure file itself is preserved on disk under
% figures/10bar_failure_manifold.pdf and the original block is kept
% verbatim below so the change is fully reversible. The disclosure
% is recorded in cover-letter.tex.
% \begin{figure}
%     \centering
%     \includegraphics[width=\columnwidth]{figures/10bar_failure_manifold.pdf}
%     \caption{The failure probabilities and constraint violations of different manifold learning methods for the ten-beam optimization problem.} 
%     \label{fig:tenbar_compre}
% \end{figure}

\begin{figure}
    \centering
    \includegraphics[width=\columnwidth]{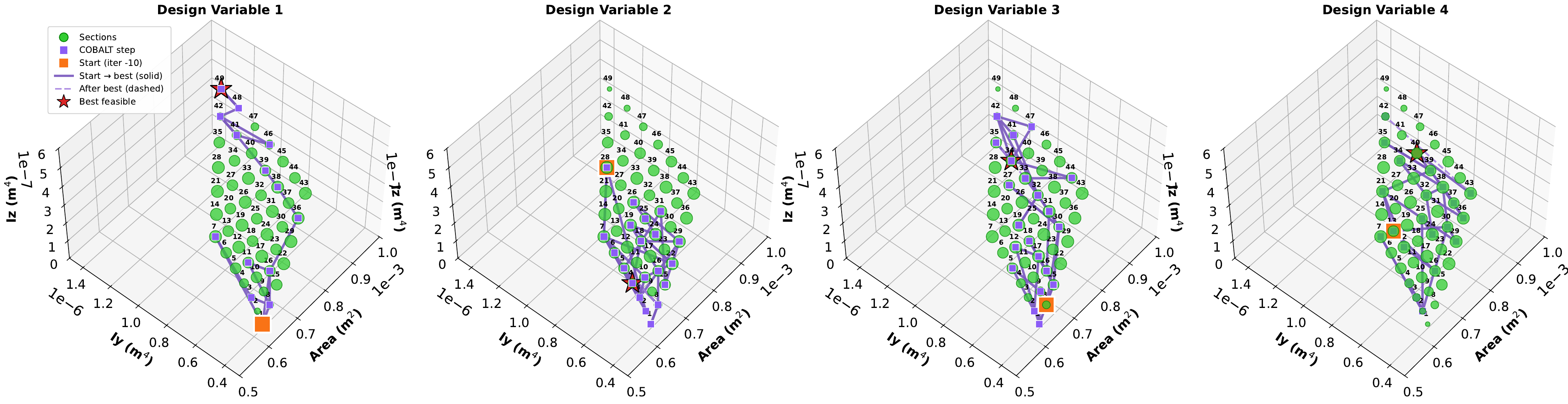}
    \caption{The optimization path in the original physical attribute space for the ten-beam optimization problem.}
    \label{fig:tenbar_physical_path}
\end{figure}

\begin{figure}
    \centering
    \includegraphics[width=\columnwidth]{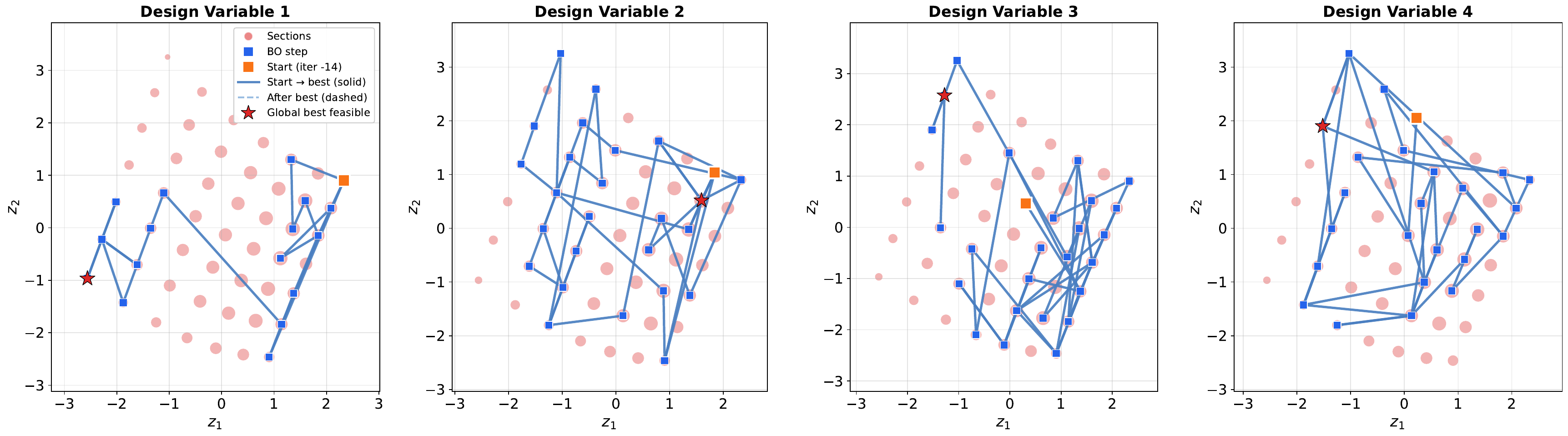}
    \caption{The optimization path in the low-dimensional design space with uncertainties for the ten-beam optimization problem.} 
    \label{fig:tenbar_decoding}
\end{figure}

Fig.~\ref{fig:tenbar_physical_path} shows the COBALT trajectory in the original attribute space $(A, I_y, I_z)$, advancing from the initial anchor (solid square) to the best feasible design (star) via catalog-valid discrete steps. 
Fig.~\ref{fig:tenbar_decoding} depicts the same trajectory in the low-dimensional latent space, with pink circles marking catalog anchors and blue squares the BO-evaluated steps. 
The solid blue path goes from the initial anchor (orange square) to the best feasible design (red star), and the dashed segment records subsequent exploration.
Every evaluated candidate falls on a catalog anchor, confirming that the graph-based search operates exclusively on physically valid discrete instances.

\subsection{The spatial dome structure}
\label{sc4_2}
\begin{figure}
    \centering
    \includegraphics[width=0.5\columnwidth]{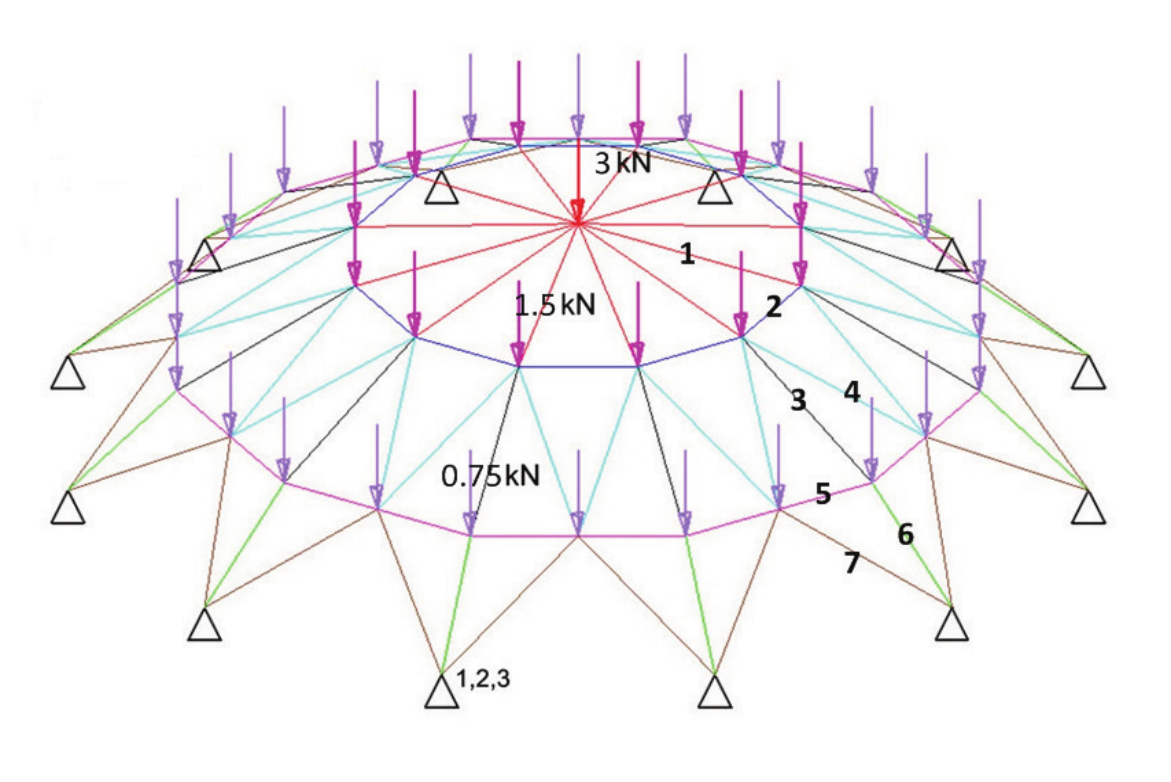}
    \caption{The load-case illustration of the dome structure.} 
    \label{fig:dome}
\end{figure}

The second benchmark is a 120-beam spatial dome with 49 nodes and 120 three-dimensional beam elements. As shown in Fig.~\ref{fig:dome}, the 120 beams are partitioned into seven symmetry-preserving design groups, giving seven categorical variables. Each group draws a common cross-section from a catalog of $n=49$ rectangular hollow sections. The sections are parameterized by outer dimensions $(L,H)$ at fixed wall thickness $t=4\,\mathrm{mm}$, and stored in the catalog as $(A,I_y,I_z)$. The characteristic span (maximum footprint diameter) is $L_{\mathrm{char}}=2l_{3}=31.78\,\mathrm{m}$. The external loads are shown in Fig.~\ref{fig:dome}, with gravity also applied.

The robust optimization problem is formulated analogously to Eq. \eqref{tenbar_obj}, with additional consideration of local buckling constraints:
\begin{equation}
\label{spatial_dome_obj}
\begin{split}
&\text{min.:}\hspace{5mm}\mathcal{J}_{\texttt{robust}}(\textbf{x}_{\mbox{\tiny{1}}},\textbf{x}_{\mbox{\tiny{2}}},\cdots,\textbf{x}_{\mbox{\tiny{7}}})=\mathbb{E}_{\boldsymbol{\xi}}\left[\frac{1}{2}\textbf{u}^{\mbox{\tiny{T}}}\textbf{K}\textbf{u}\right] + \gamma \sqrt{\mathbb{V}_{\boldsymbol{\xi}}\left[\frac{1}{2}\textbf{u}^{\mbox{\tiny{T}}}\textbf{K}\textbf{u}\right]};\\
&\text{s.}\hspace{1.5mm}\text{t.:}\hspace{5.8mm}mass(\textbf{x}_{\mbox{\tiny{1}}},\textbf{x}_{\mbox{\tiny{2}}},\cdots,\textbf{x}_{\mbox{\tiny{7}}})-4000\leq 0;\\
&\hspace{13mm}\textbf{K}\textbf{u}=\textbf{P};\\
&\hspace{13mm}\max(\textbf{F}-\textbf{F}_y^{\mbox{\tiny{cr}}})\leq 0,\hspace{2mm}\textbf{F}_y^{\mbox{\tiny{cr}}}=(f_y^{\mbox{\tiny{cr1}}},f_y^{\mbox{\tiny{cr2}}},\cdots,f_y^{\mbox{\tiny{cr120}}});\\
&\hspace{13mm}\max(\textbf{F}-\textbf{F}_z^{\mbox{\tiny{cr}}})\leq 0,\hspace{2mm}\textbf{F}_z^{\mbox{\tiny{cr}}}=(f_z^{\mbox{\tiny{cr1}}},f_z^{\mbox{\tiny{cr2}}},\cdots,f_z^{\mbox{\tiny{cr120}}});\\
&\hspace{13mm}\textbf{x}_{\mbox{\tiny{i}}}\in \{\textbf{x}^{\mbox{\tiny{1}}}_{\mbox{\tiny{i}}},\textbf{x}^{\mbox{\tiny{2}}}_{\mbox{\tiny{i}}},\cdots,\textbf{x}^{\mbox{\tiny{49}}}_{\mbox{\tiny{i}}}\},\hspace{2mm}\mbox{\normalsize{i}}=1,2,\cdots,7;\\
&\hspace{13mm}\textbf{x}^{\mbox{\tiny{j}}}_{\mbox{\tiny{i}}}=(A^{\mbox{\tiny{j}}}_{\mbox{\tiny{i}}},I_{y}{}^{\mbox{\tiny{j}}}_{\mbox{\tiny{i}}},I_{z}{}^{\mbox{\tiny{j}}}_{\mbox{\tiny{i}}})^{\mbox{\tiny{T}}},\hspace{2mm}\mbox{\normalsize{j}}=1,2,\cdots,49.\\
\end{split}
\end{equation}


\begin{figure}
    \centering
    \includegraphics[width=\columnwidth]{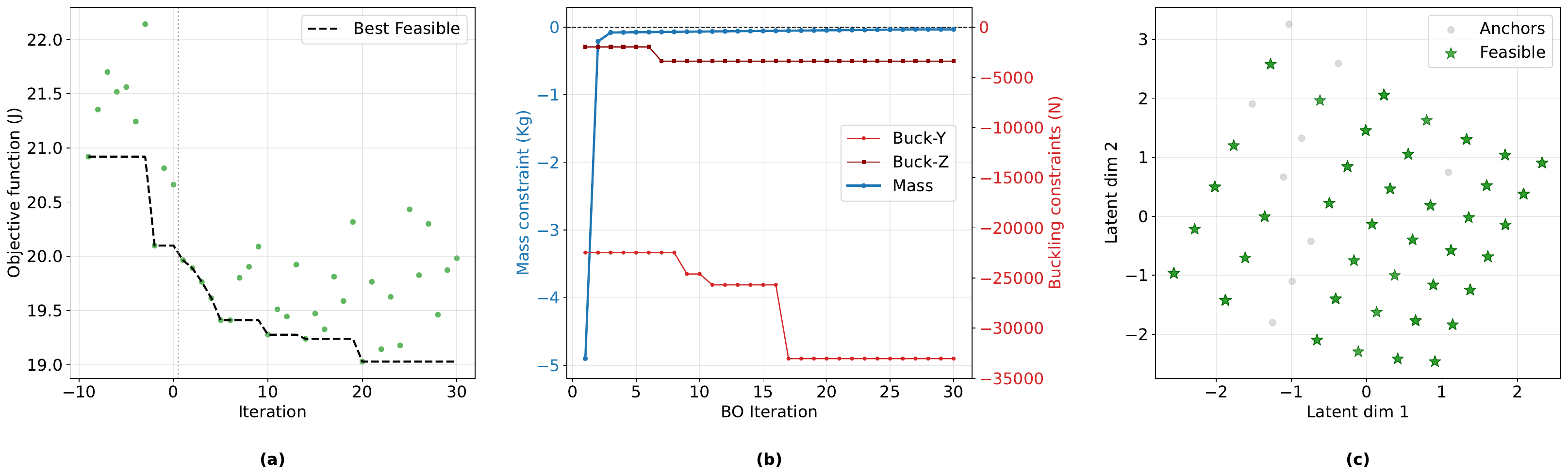}
    \caption{The convergence history of the dome optimization: (a) robust objective convergence, (b) mass and buckling-constraint evolution, and (c) feasible anchored designs in the latent space.}
    \label{fig:dome_convergence}
\end{figure}

\begin{figure}
    \centering
    \includegraphics[width=\columnwidth]{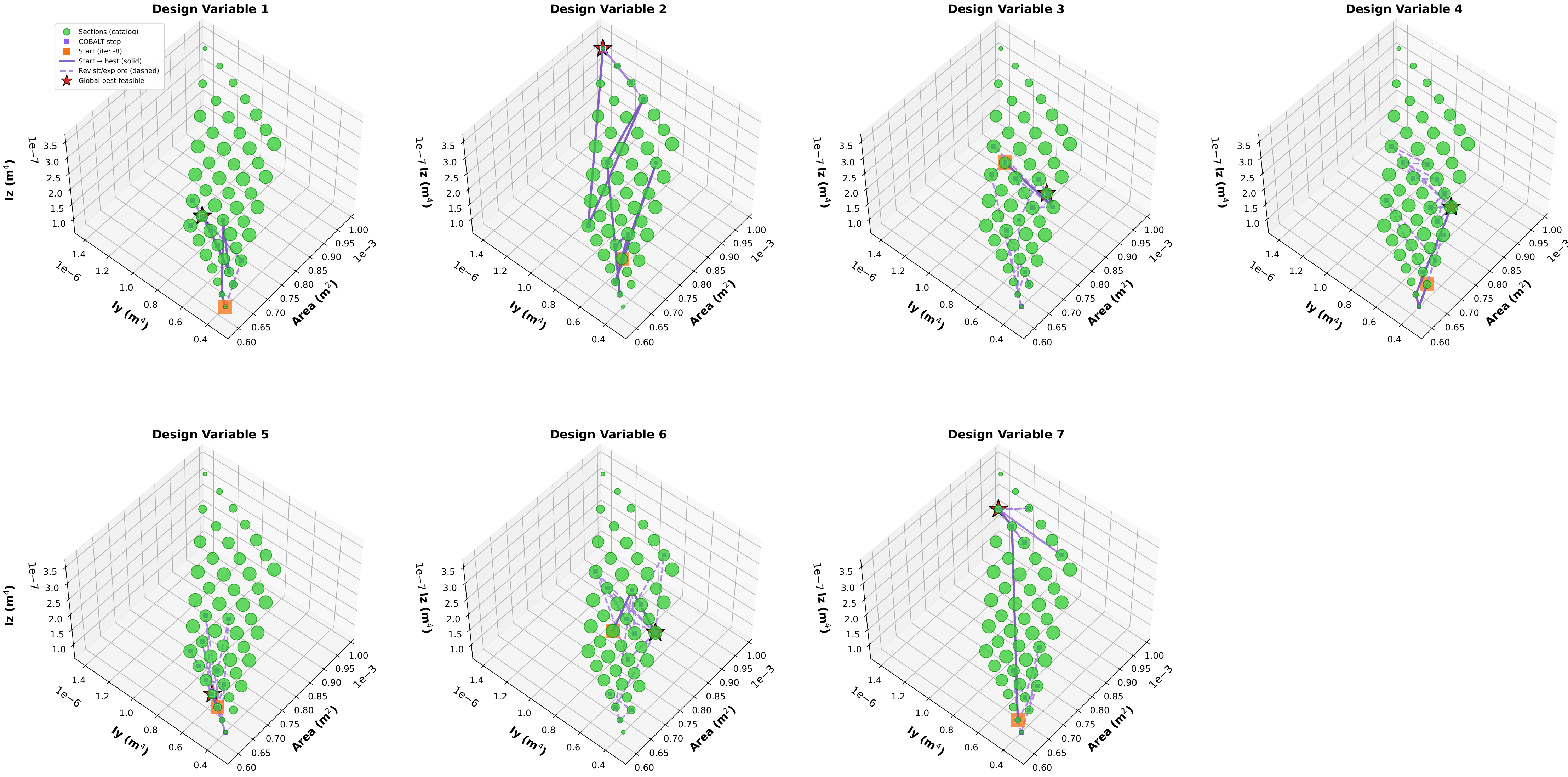}
    \caption{The optimization path in the original physical attribute space for the dome optimization problem.}
    \label{fig:dome_physical_path}
\end{figure}

\begin{figure}
    \centering
    \includegraphics[width=\columnwidth]{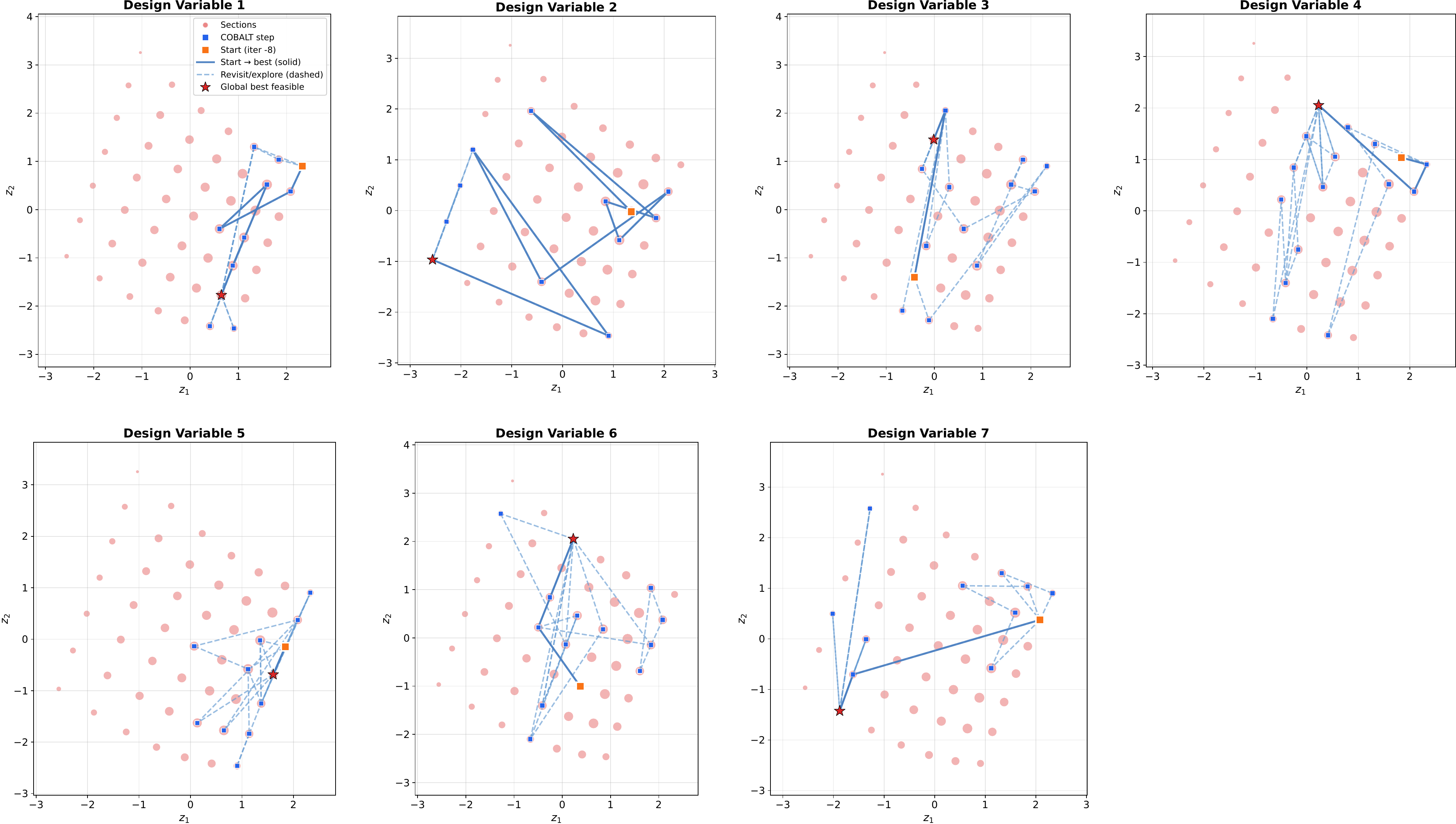}
    \caption{The optimization path in the low-dimensional design space with uncertainties for the dome optimization problem.}
    \label{fig:dome_latent_path}
\end{figure}1

For the spatial dome, the uncertain parameter vector $\boldsymbol{\xi}$ augments the categorical section-attribute uncertainties inherited from the ten-beam benchmark with three additional physical sources, yielding four independent uncertainty channels:
\begin{itemize}
    \item Section attributes $(A, I_y, I_z)$: independent log-normal multipliers, unit mean, $\mathrm{CoV}=0.05$.
    \item Young's modulus: $\widetilde{E}=E_0\eta_E$, $\eta_E$ log-normal with unit mean and $\mathrm{CoV}=0.05$; $E_0=2.1\times10^{11}\,\mathrm{Pa}$.
    \item External loads: $\widetilde{P}_i=P_{i,0}\eta_P$, with a shared log-normal $\eta_P$ (unit mean, $\mathrm{CoV}=0.01$). Nominal $-z$ forces: $P_{\mathrm{apex}}=3000\,\mathrm{N}$, $P_{\mathrm{mid}}=1500\,\mathrm{N}$ (12 intermediate-ring nodes), $P_{\mathrm{outer}}=750\,\mathrm{N}$ (24 outer-ring nodes).
    \item Geometric imperfections: $\Delta\boldsymbol{r}_{k}\sim\mathcal{N}(\mathbf{0},(0.001L_{\mathrm{char}})^{2}\mathbf{I}_{3})$ across all Cartesian components of the 49 nodes, with $L_{\mathrm{char}}=31.78\,\mathrm{m}$, giving $\sigma_{\mathrm{geom}}=31.78\,\mathrm{mm}$ per component, consistent with fabrication and erection tolerances.
\end{itemize}
The four uncertainty sources are sampled independently at every Monte Carlo realization. The robust metrics are estimated via MC-FEA with $N_{MC} = 100$ samples per candidate design, introducing heteroscedastic observation noise as modeled in Eq. \eqref{eq_noise}.

Fig. \ref{fig:dome_convergence} presents the convergence histories of the robust objective, the mass constraint, and the buckling constraints over 200 MC-FEA evaluations for the 120-beam dome. The performance gap between COBALT and the baseline methods is substantially amplified compared to the ten-beam case, reflecting the increased difficulty of navigating a higher-dimensional combinatorial space under uncertainties.

In the reported run, COBALT attains the lowest verified feasible robust objective among the considered baselines.
% COBALT achieves the best feasible robust objective among all compared methods within the allocated evaluation budget. 
The deterministic manifold optimization method fails to locate competitive feasible designs within the same budget, illustrating the difficulty of optimizing over a continuous manifold in high-dimensional categorical OUU.
This behavior is consistent with the decoding sensitivity expected in high-dimensional categorical spaces.
As the number of categorical variables increases, the probability of a continuous pseudo-optimum mapping to a topologically distant discrete anchor grows, leading to structurally inferior or constraint-violating configurations. In contrast, COBALT's discrete graph acquisition ensures that every recommended configuration is physically admissible, supporting stable improvement throughout the optimization campaign.

Fig. \ref{fig:dome_physical_path} illustrates the optimization trajectory of COBALT in the original physical attribute space, where each of the seven design variables is represented by its cross-sectional attributes $(A, I_y, I_z)$. The search path progresses from the initial anchor toward the best feasible design through a sequence of catalog-valid discrete steps, demonstrating the structured exploration of the seven-dimensional combinatorial design space.

Fig. \ref{fig:dome_latent_path} further visualizes the corresponding optimization trajectory in the low-dimensional latent space under uncertainties. The discrete catalog anchors are fixed points in the latent manifold, and the blue squares indicate successive BO-evaluated steps. The solid path traces the progression from the initial anchor (orange square) to the best observed feasible design (red star). The dashed segment records the exploratory steps taken after the current best was identified. Every evaluated candidate coincides with a catalog anchor, confirming that the graph-based search operates exclusively on physically valid discrete instances throughout the optimization.

\subsection{The planar 105-beam structure}
\label{sc4_4}
\begin{figure}
    \centering
    \includegraphics[width=0.5\columnwidth]{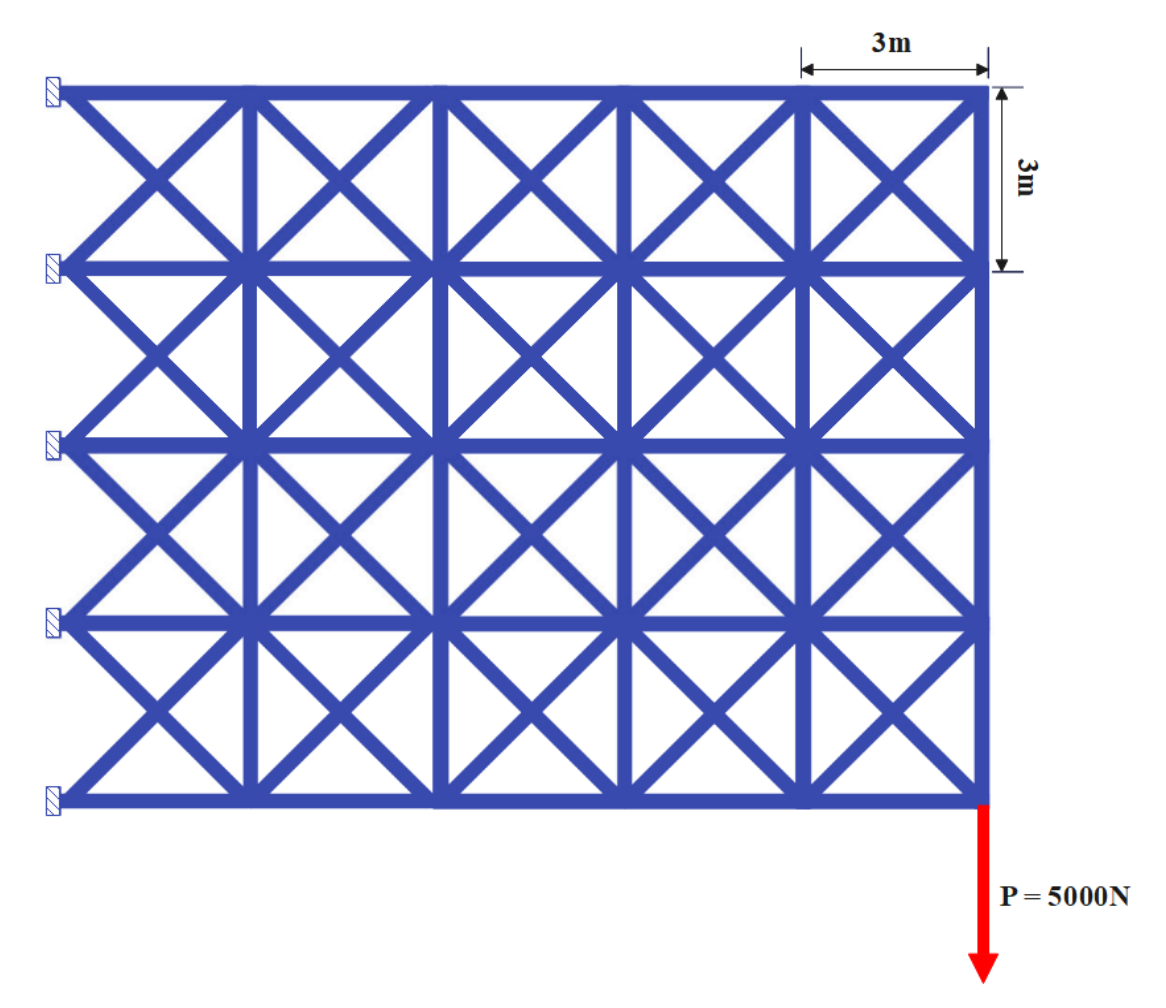}
    \caption{The load-case illustration of the 105-beam structure.}
    \label{fig:105beam}
\end{figure}

As shown in Fig. \ref{fig:105beam}, the 105 beams act as independent categorical variables, each drawn from the same 54-profile catalog. The lower-left end of the strucutre carries a concentrated force $P=5000\,\mathrm{N}$ together with gravity.

The robust optimization problem is formulated as:
\begin{equation}
\label{105bar_obj}
\begin{split}
&\text{min.:}\hspace{5mm}\mathcal{J}_{\texttt{robust}}(\textbf{x}_{\mbox{\tiny{1}}},\textbf{x}_{\mbox{\tiny{2}}},\cdots,\textbf{x}_{\mbox{\tiny{105}}})=\mathbb{E}_{\boldsymbol{\xi}}\left[\frac{1}{2}\textbf{u}^{\mbox{\tiny{T}}}\textbf{K}\textbf{u}\right] + \gamma \sqrt{\mathbb{V}_{\boldsymbol{\xi}}\left[\frac{1}{2}\textbf{u}^{\mbox{\tiny{T}}}\textbf{K}\textbf{u}\right]};\\
&\text{s.}\hspace{1.5mm}\text{t.:}\hspace{5.8mm}mass(\textbf{x}_{\mbox{\tiny{1}}},\textbf{x}_{\mbox{\tiny{2}}},\cdots,\textbf{x}_{\mbox{\tiny{105}}})-5000\leq 0;\\
&\hspace{13mm}\textbf{K}\textbf{u}=\textbf{P};\\
&\hspace{13mm}\max(\textbf{F}-\textbf{F}_y^{\mbox{\tiny{cr}}})\leq 0,\hspace{2mm}\textbf{F}_y^{\mbox{\tiny{cr}}}=(f_y^{\mbox{\tiny{cr1}}},f_y^{\mbox{\tiny{cr2}}},\cdots,f_y^{\mbox{\tiny{cr105}}});\\
&\hspace{13mm}\max(\textbf{F}-\textbf{F}_z^{\mbox{\tiny{cr}}})\leq 0,\hspace{2mm}\textbf{F}_z^{\mbox{\tiny{cr}}}=(f_z^{\mbox{\tiny{cr1}}},f_z^{\mbox{\tiny{cr2}}},\cdots,f_z^{\mbox{\tiny{cr105}}});\\
&\hspace{13mm}\textbf{x}_{\mbox{\tiny{i}}}\in \{\textbf{x}^{\mbox{\tiny{1}}}_{\mbox{\tiny{i}}},\textbf{x}^{\mbox{\tiny{2}}}_{\mbox{\tiny{i}}},\cdots,\textbf{x}^{\mbox{\tiny{49}}}_{\mbox{\tiny{i}}}\},\hspace{2mm}\mbox{\normalsize{i}}=1,2,\cdots,105;\\
&\hspace{13mm}\textbf{x}^{\mbox{\tiny{j}}}_{\mbox{\tiny{i}}}=(A^{\mbox{\tiny{j}}}_{\mbox{\tiny{i}}},I_{y}{}^{\mbox{\tiny{j}}}_{\mbox{\tiny{i}}},I_{z}{}^{\mbox{\tiny{j}}}_{\mbox{\tiny{i}}})^{\mbox{\tiny{T}}},\hspace{2mm}\mbox{\normalsize{j}}=1,2,\cdots,49.\\
\end{split}
\end{equation}

The uncertainties model is consistent with the preceding examples:
\begin{itemize}
    \item Section attributes $(A, I_y, I_z)$: independent log-normal multipliers, unit mean, $\mathrm{CoV}=0.05$.
    \item Young's modulus: $\widetilde{E}=E_0\eta_E$, $\eta_E$ log-normal, unit mean, $\mathrm{CoV}=0.05$.
    \item External loads: $\widetilde{P}_i=P_{i,0}\eta_{P,i}$, $\eta_{P,i}$ log-normal, unit mean, $\mathrm{CoV}=0.01$.
\end{itemize}

The MC-FEA uses $N_{MC} = 500$ samples per evaluation. The total combinatorial search space is $54^{105}$, a prohibitively large number that renders any exhaustive strategy infeasible.

With 105 independent categorical variables, this benchmark is the first medium-scale stress test for COBALT and the regime where the random tree decomposition matters. With no grouping symmetry, the interaction topology among variables is unknown a priori. COBALT's uniformly random spanning-tree sampling exposes different pairwise couplings at each iteration, keeping the surrogate from committing to one fragile interaction pattern. Within the same budget, deterministic manifold optimization fails to find competitive feasible designs at this scale, since the feasible-decoding fraction collapses rapidly with dimensionality. Continuous latent-space Bayesian optimization still finds reasonable designs early on, but its decoding failure rate climbs sharply. At 105 variables, a single rounding-off error in one variable can cascade through the truss connectivity and breach buckling constraints in distant members. COBALT sidesteps this mode by confining every acquisition step to the discrete anchored graph. Its SAAS-GP prior also masks irrelevant latent dimensions, preserving surrogate fidelity under the tight budget.

\begin{figure}
    \centering
    \includegraphics[width=\columnwidth]{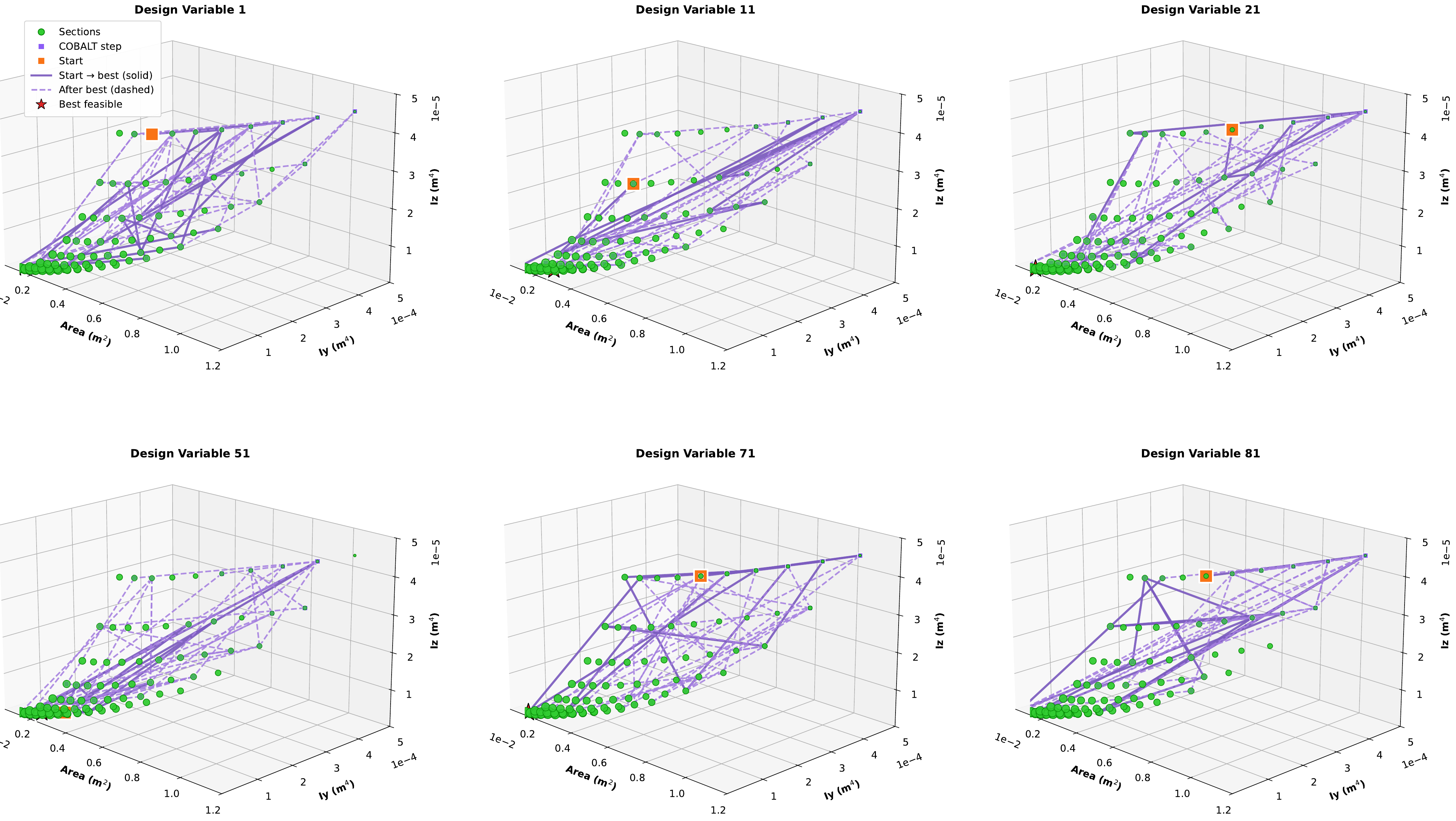}
    \caption{The optimization path in the original physical attribute space for the 105-beam optimization problem.}
    \label{fig:105bar_physical_path}
\end{figure}

\begin{figure}
    \centering
    \includegraphics[width=\columnwidth]{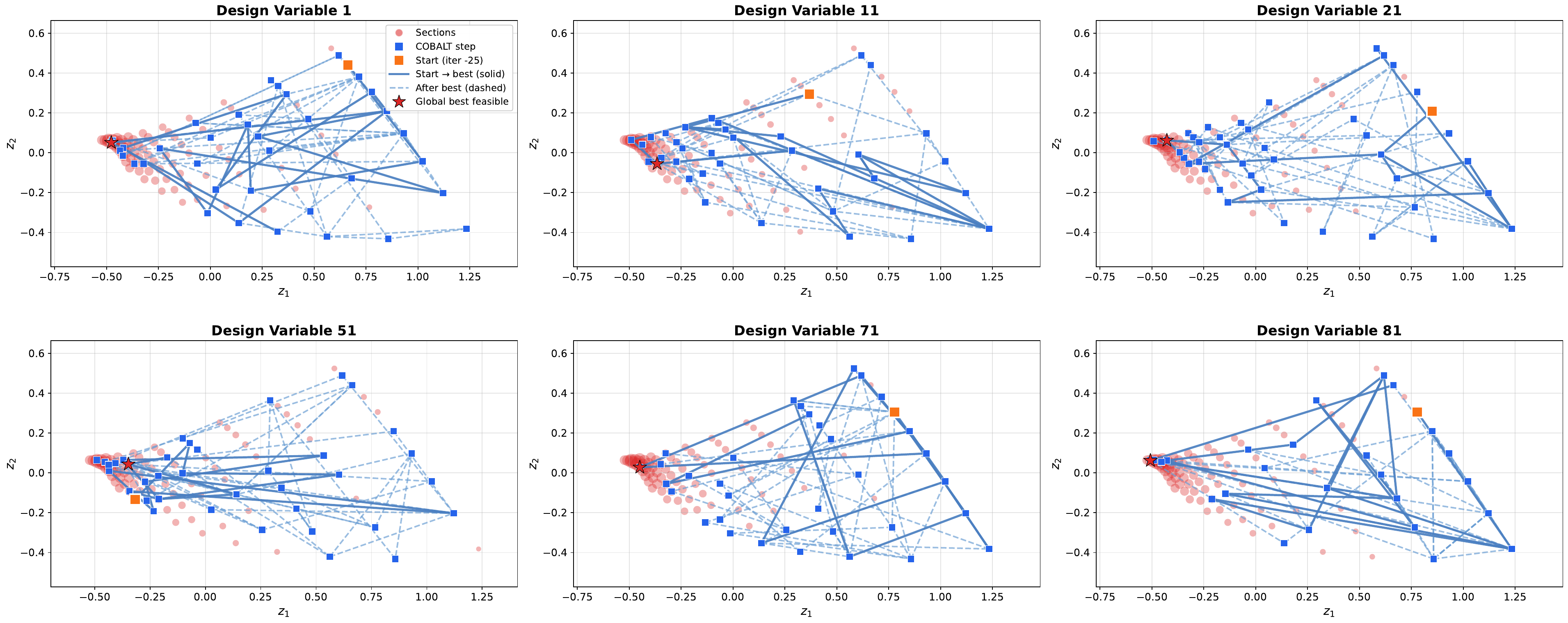}
    \caption{The optimization path in the low-dimensional design space with uncertainties for the 105-beam optimization problem.}
    \label{fig:105bar_latent_path}
\end{figure}

Fig.~\ref{fig:105bar_physical_path} shows a representative optimization trajectory in the physical attribute space. The plotted variables are represented by their cross-sectional attributes $(A, I_y, I_z)$. COBALT first locates a coarse allocation of heavy and light sections across the structure. It then refines individual member assignments, reflecting the alternation between trust-region expansion for global exploration and graph-based discrete search for local exploitation. Fig.~\ref{fig:105bar_latent_path} visualizes the same trajectory in the low-dimensional latent space. Catalog anchors appear as fixed points and blue squares mark successive BO-evaluated steps. The solid path connects the initial anchor (orange square) to the best observed feasible design (red star). The dashed segment records the exploratory steps taken after this best design was identified. Every evaluated candidate coincides with a catalog anchor, confirming that the search operates only on physically valid discrete instances.

\subsection{The high-dimensional 1564-beam structure}
\label{sc4_5}
The fifth benchmark is a high-dimensional spatial 1564-beam structure, representing the largest categorical optimization problem considered in this study. The 1564 beams are treated as independent categorical design variables, each selected from the same catalog of 49 standard steel profiles with attributes $(A, I_y, I_z, J_x)$. This example is designed to evaluate COBALT's scalability in handling an extremely large combinatorial space under uncertainties. The structure bears the dead load of the beams and a heavy snow load on the ceiling of the structure ($200\,\mathrm{N/m^2}$).

The robust optimization problem is formulated as:
\begin{equation}
    \label{beam1564_obj}
    \begin{split}
&\text{min.:}\hspace{5mm}\mathcal{J}_{\texttt{robust}}(\textbf{x}_{\mbox{\tiny{1}}},\textbf{x}_{\mbox{\tiny{2}}},\cdots,\textbf{x}_{\mbox{\tiny{1564}}})=\mathbb{E}_{\boldsymbol{\xi}}\left[\frac{1}{2}\textbf{u}^{\mbox{\tiny{T}}}\textbf{K}\textbf{u}\right] + \gamma \sqrt{\mathbb{V}_{\boldsymbol{\xi}}\left[\frac{1}{2}\textbf{u}^{\mbox{\tiny{T}}}\textbf{K}\textbf{u}\right]};\\
&\text{s.}\hspace{1.5mm}\text{t.:}\hspace{5.8mm}mass(\textbf{x}_{\mbox{\tiny{1}}},\textbf{x}_{\mbox{\tiny{2}}},\cdots,\textbf{x}_{\mbox{\tiny{1564}}})-70000\leq 0;\\
&\hspace{13mm}\textbf{K}\textbf{u}=\textbf{P};\\
&\hspace{13mm}\max(\textbf{F}-\textbf{F}_y^{\mbox{\tiny{cr}}})\leq 0,\hspace{2mm}\textbf{F}_y^{\mbox{\tiny{cr}}}=(f_y^{\mbox{\tiny{cr1}}},f_y^{\mbox{\tiny{cr2}}},\cdots,f_y^{\mbox{\tiny{cr1564}}});\\
&\hspace{13mm}\max(\textbf{F}-\textbf{F}_z^{\mbox{\tiny{cr}}})\leq 0,\hspace{2mm}\textbf{F}_z^{\mbox{\tiny{cr}}}=(f_z^{\mbox{\tiny{cr1}}},f_z^{\mbox{\tiny{cr2}}},\cdots,f_z^{\mbox{\tiny{cr1564}}});\\
&\hspace{13mm}\textbf{x}_{\mbox{\tiny{i}}}\in \{\textbf{x}^{\mbox{\tiny{1}}}_{\mbox{\tiny{i}}},\textbf{x}^{\mbox{\tiny{2}}}_{\mbox{\tiny{i}}},\cdots,\textbf{x}^{\mbox{\tiny{49}}}_{\mbox{\tiny{i}}}\},\hspace{2mm}\mbox{\normalsize{i}}=1,2,\cdots,1564;\\
&\hspace{13mm}\textbf{x}^{\mbox{\tiny{j}}}_{\mbox{\tiny{i}}}=(A^{\mbox{\tiny{j}}}_{\mbox{\tiny{i}}},I_{y}{}^{\mbox{\tiny{j}}}_{\mbox{\tiny{i}}},I_{z}{}^{\mbox{\tiny{j}}}_{\mbox{\tiny{i}}},J_{x}{}^{\mbox{\tiny{j}}}_{\mbox{\tiny{i}}})^{\mbox{\tiny{T}}},\hspace{2mm}\mbox{\normalsize{j}}=1,2,\cdots,49.\\
\end{split}
\end{equation}

\begin{figure}
    \centering
    \includegraphics[width=0.75\columnwidth]{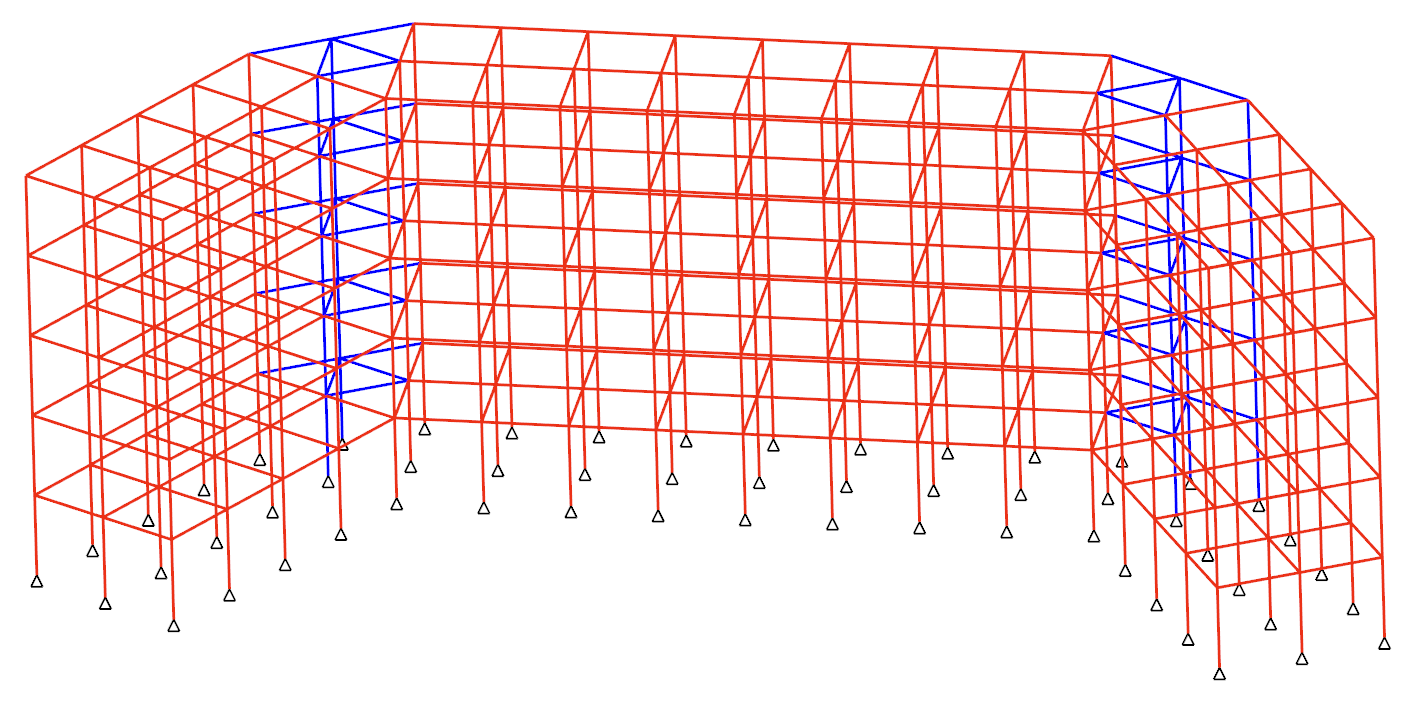}
    \caption{The load-case illustration of the high-dimensional 1564-beam structure.}
    \label{fig:1564beam}
\end{figure}

The uncertainties model includes:
\begin{itemize}
    \item Section attributes $(A, I_y, I_z, J_x)$: independent log-normal multipliers, unit mean, $\mathrm{CoV}=0.05$.
    \item Young's modulus: $\widetilde{E}=E_0\eta_E$, $\eta_E$ log-normal, unit mean, $\mathrm{CoV}=0.05$.
    \item External loads (snow): $\widetilde{P}_i=P_{i,0}\eta_{P,i}$, $\eta_{P,i}$ log-normal, unit mean, $\mathrm{CoV}=0.01$.
    \item Geometric imperfections: nodal-coordinate perturbations $\Delta\boldsymbol{r}_k \sim \mathcal{N}(\mathbf{0},(0.001L_{\mathrm{char}})^2\mathbf{I}_3)$, with $L_{\mathrm{char}}$ the characteristic span.
\end{itemize}
The MC-FEA uses $N_{MC} = 500$ samples per evaluation. The total combinatorial search space is $54^{1564}$, a number that exceeds any practical enumeration by many orders of magnitude.

This example is the largest scalability test for COBALT. At 1564 independent categorical variables, the search-space dimensionality far exceeds the regime where conventional surrogate-assisted methods stay effective without extra structure. Two COBALT mechanisms are critical at this scale. First, the SAAS horseshoe prior (Eq.~\eqref{horseshoe}) shrinks the inverse lengthscales of less influential latent dimensions, letting the Additive SAAS-GP focus on members that dominate the robust objective. Second, the random tree decomposition caps each surrogate component at two variables, keeping the additive surrogate tractable even with $D=m\times 1564$. Deterministic manifold optimization yields infeasible or low-quality designs across the budget, illustrating the difficulty of continuous optimization on a mapped catalog manifold at this scale. Continuous latent-space Bayesian optimization suffers frequent decoding failure: with 1564 variables, rounding rarely recovers the intended discrete configuration. COBALT avoids this mode by construction and keeps improving via discrete graph acquisition.

\begin{figure}
    \centering
    \includegraphics[width=\columnwidth]{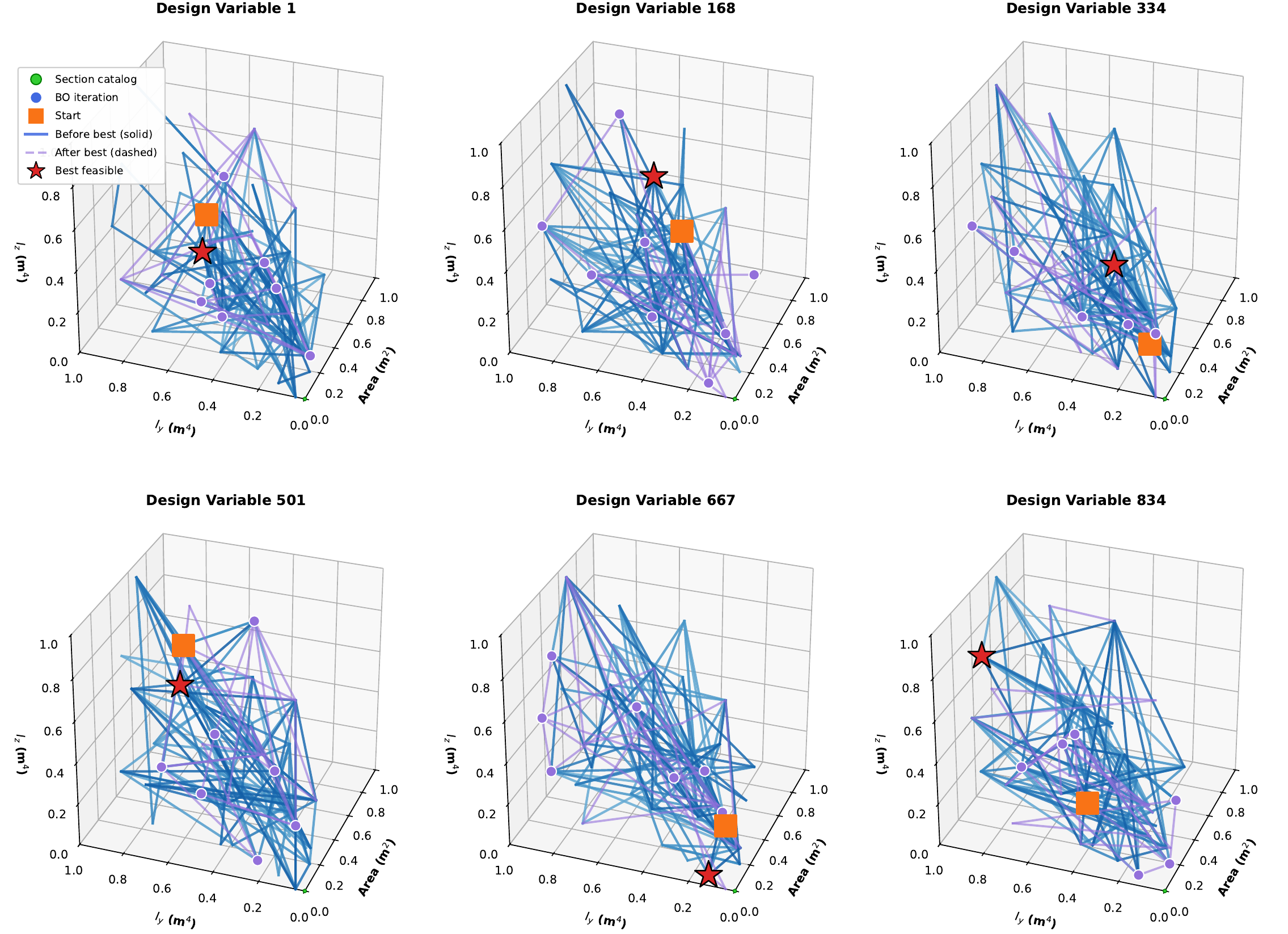}
    \caption{The optimization path in the original physical attribute space for the 1564-beam optimization problem.}
    \label{fig:1564beam_physical_path}
\end{figure}

\begin{figure}
    \centering
    \includegraphics[width=\columnwidth]{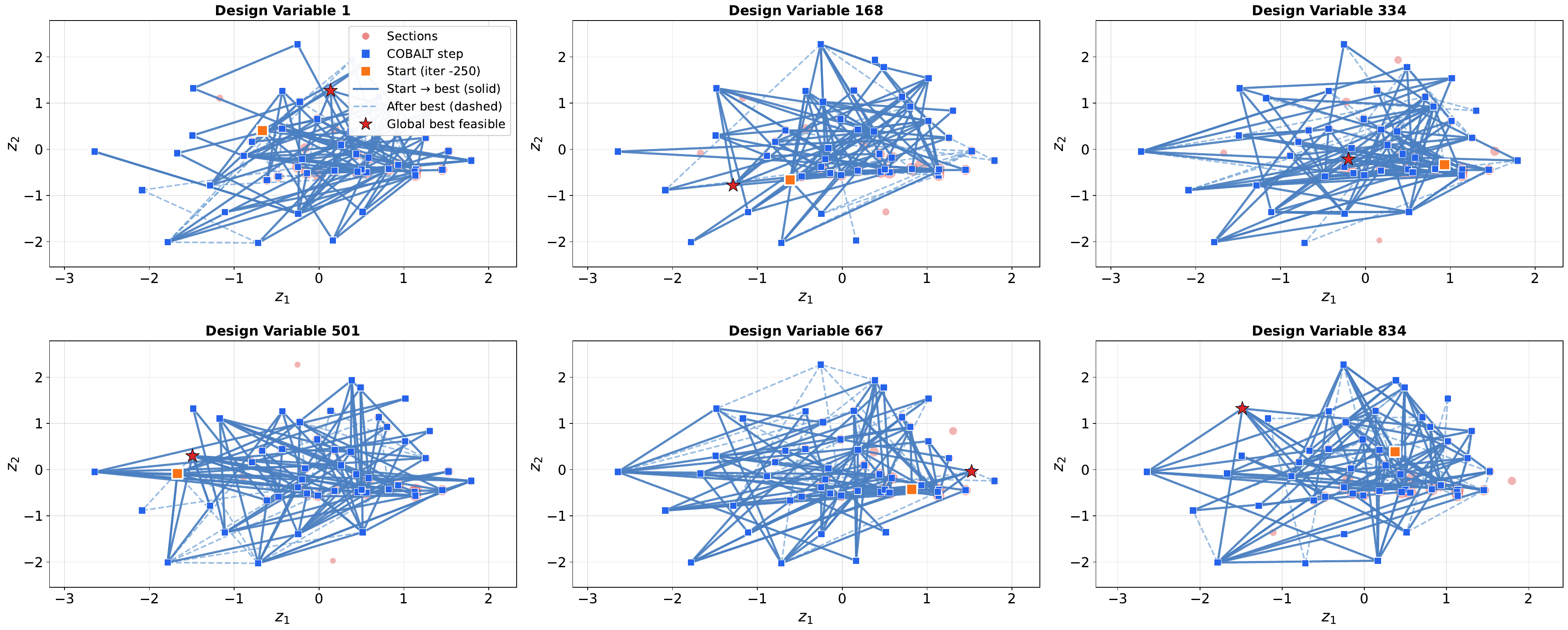}
    \caption{The optimization path in the low-dimensional design space with uncertainties for the 1564-beam optimization problem.}
    \label{fig:1564beam_latent_path}
\end{figure}

Fig.~\ref{fig:1564beam_physical_path} shows representative optimization trajectories in the original physical-attribute space. The plotted variables are the cross-sectional attributes $(A, I_y, I_z, J_x)$. Despite the extreme dimensionality, the trajectory progresses coherently from the initial random assignment toward a structurally efficient design. The SAAS-GP identifies load-critical members early and assigns them heavier sections. Peripheral members receive lighter profiles, reflecting the surrogate's automatic relevance determination. Fig.~\ref{fig:1564beam_latent_path} visualizes the same trajectory in the low-dimensional latent space under uncertainties. The catalog anchors are fixed in the latent manifold and the blue squares mark successive BO steps. The solid path traces the run from the initial anchor (orange square) to the best feasible design (red star). The dashed segment records exploratory steps taken after the current best was found. Every evaluated candidate coincides with a catalog anchor, confirming that the graph-based search operates only on physically valid discrete instances.

\section{Conclusions}
\label{sc5}
This paper presents COBALT, a Bayesian optimization framework for robust categorical structural design under aleatoric uncertainties that addresses two coupled difficulties: preserving catalog admissibility after dimensionality reduction and curbing the number of expensive uncertainty-aware finite-element evaluations in large combinatorial design spaces. COBALT locks the dimensionality-reduced catalog as a fixed discrete anchor set rather than a continuous domain, so the acquisition search stays on admissible catalog configurations and avoids the rounding-off step of continuous latent-space methods; Sparse Axis-Aligned Subspace (SAAS) priors in the Gaussian Process surrogate yield a sparse fully Bayesian model for noisy MC-FEA observations under tight evaluation budgets; and Dijkstra-based graph operators carry out acquisition within latent trust regions at overhead negligible against the dominant MC-FEA oracle cost. On the ten-beam planar truss, the 120-beam spatial dome, the planar 105-beam structure, and the 1564-beam structure, COBALT delivers feasible catalog designs under the prescribed budgets and matches or beats the continuous latent-space and deterministic manifold-optimization baselines, with evidence limited to the reported benchmarks and run configurations. Limitations include the pre-optimization-fixed Isomap embedding, which may need adaptation for non-stationary catalogs, and the common-catalog assumption; heterogeneous and mixed categorical--continuous catalogs, repeated trials with confidence intervals and feasibility-rate summaries, and multi-fidelity uncertainty evaluation are promising directions to broaden applicability and cut MC-FEA cost.

\section*{Acknowledgment}
The authors gratefully acknowledge the financial support from the National Natural Science Foundation of China (12572138, 12272302).

% \appendix
% \section{My Appendix}
% Appendix sections are coded under 

% \printcredits

%% Loading bibliography style file
% Numbered citation style (matches the Engineering Structures requirement
% and the original Article.tex configuration). The cas-model2-names style
% renders citations as author-year, which is why citations were not
% appearing as bracketed numbers; switch to the numbered Elsevier style.
%\bibliographystyle{cas-model2-names}
\bibliographystyle{elsarticle-num-names}

% Loading bibliography database
\bibliography{cas-refs}

%\vskip3pt

% \bio{figs/cas-pic1}
% Author biography with author photo.
% Author biography. Author biography. Author biography.
% \endbio

% \vskip3pc

% \bio{figs/cas-pic1}
% Author biography with author photo.
% Author biography. Author biography. Author biography.
% \endbio

\end{document}